\documentclass[10pt,twocolumn,letterpaper]{article}

\usepackage[pagenumbers]{cvpr} 
\usepackage[accsupp]{axessibility} 

\definecolor{cvprblue}{rgb}{0.21,0.49,0.74}
\usepackage[pagebackref,breaklinks,colorlinks,allcolors=cvprblue]{hyperref}
\usepackage{multirow}
\usepackage{array}
\usepackage{colortbl}
\usepackage[accsupp]{axessibility}

\definecolor{mygray}{HTML}{F5F5F5}
\definecolor{palegreen}{HTML}{EAF4EA} 
\definecolor{palered}{HTML}{FBECEC}   
\definecolor{paleblue}{HTML}{EAF2FD}  
\def\paperID{3896} 
\def\confName{CVPR}
\def\confYear{2026}

\title{DreamShot: Personalized Storyboard Synthesis with Video Diffusion Prior}

\author{
    Junjia Huang$^{1}$\thanks{Equal Contribution.}\quad Binbin Yang$^{2*}$\thanks{Project Lead.} \quad Pengxiang Yan$^{2}$\quad Jiyang Liu$^{2}$\quad Bin Xia$^{2}$\\
    Zhao Wang$^2$\quad  Yitong Wang$^{2}$\quad  Liang Lin$^{1}$\quad  Guanbin Li$^{1}$\thanks{Corresponding Author.}\\
    {$^1$Sun Yat-sen University}, $^2$ByteDance Intelligent Creation  \\
    {\tt\small huangjj77@mail2.sysu.edu.cn, wantong1017@163.com} \\ {\tt\small linliang@ieee.org, liguanbin@mail.sysu.edu.cn}\\
    {\tt\small \{yangbinbin.3, yanpengxiang.ai, liujiyang.liu, xiabin.zj, zhaoxu.bit\}@bytedance.com}\\
    {\href{https://ll3rd.github.io/DreamShot/}{https://ll3rd.github.io/DreamShot/}}
}

\begin{document}
\twocolumn[{
  \maketitle
  \begin{center}
    \vspace{-0.7cm}
    \includegraphics[width=0.98\linewidth]{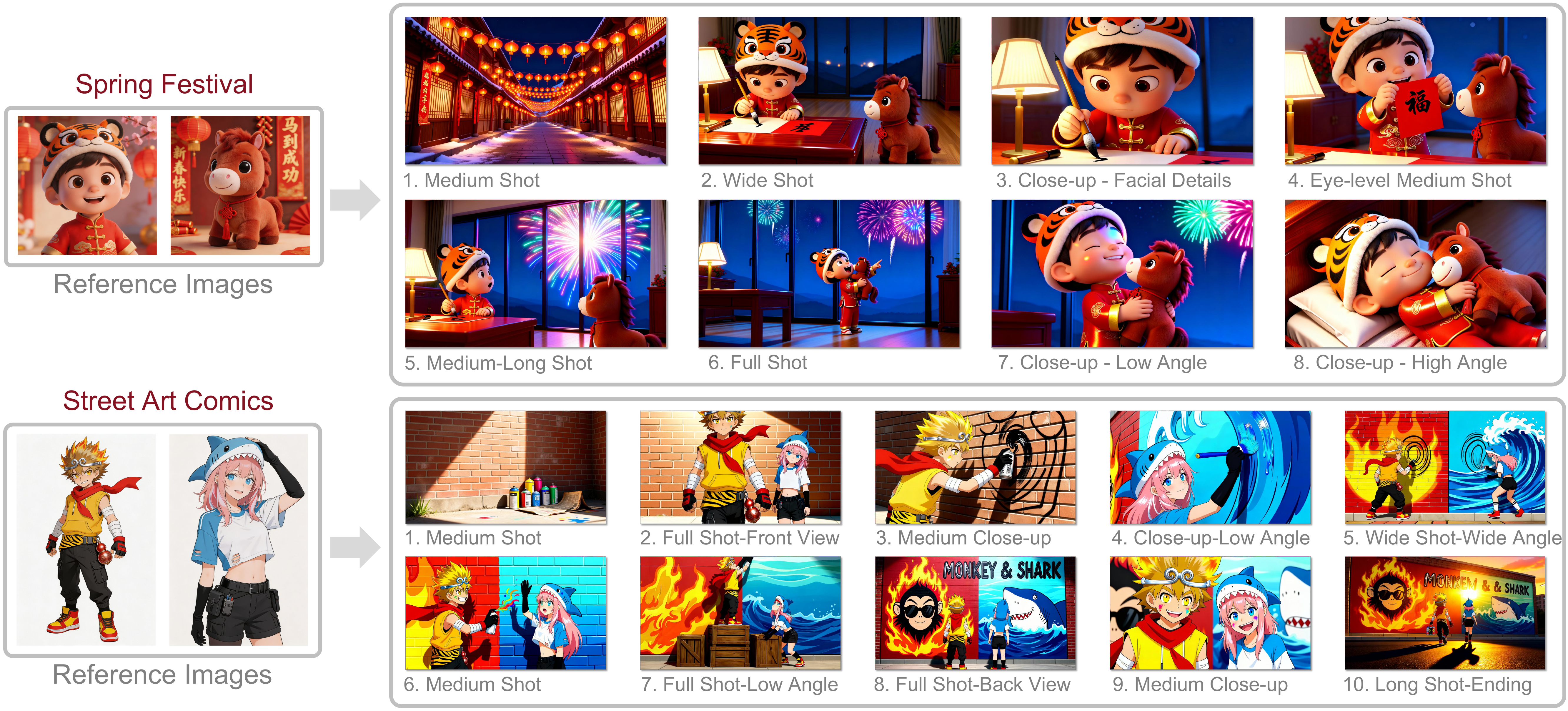}
    \vspace{-0.3cm}
    \captionof{figure}{DreamShot generates coherent, multi-shot storyboards conditioned on multiple reference images.
Given character reference images (left), DreamShot synthesizes personalized storyboard sequences (right) that preserve identity, appearance, and style across shots. The generated shots remain consistent through large viewpoint shifts, character interactions, and scene transitions, demonstrating DreamShot’s ability to maintain role consistency, scene continuity, and narrative coherence throughout multi-shot visual storytelling.}
    \label{fig: intro}
  \end{center}
}]

\maketitle

\renewcommand{\thefootnote}{\fnsymbol{footnote}} 
\footnotetext[1]{Equal Contribution. \footnotemark[2]Project Lead. \footnotemark[3]Corresponding Author.}

\begin{abstract}
Storyboard synthesis plays a crucial role in visual storytelling, aiming to generate coherent shot sequences that visually narrate cinematic events with consistent characters, scenes, and transitions. However, existing approaches are mostly adapted from text-to-image diffusion models, which struggle to maintain long-range temporal coherence, consistent character identities, and narrative flow across multiple shots. In this paper, we introduce DreamShot, a video generative model based storyboard framework that fully exploits powerful video diffusion priors for controllable multi-shot synthesis. DreamShot supports both Text-to-Shot and Reference-to-Shot generation, as well as story continuation conditioned on previous frames, enabling flexible and context-aware storyboard generation. By leveraging the spatial-temporal consistency inherent in video generative models, DreamShot produces visually and semantically coherent sequences with improved narrative fidelity and character continuity. Furthermore, DreamShot incorporates a multi-reference role conditioning module that accepts multiple character reference images and enforces identity alignment via a Role-Attention Consistency Loss, explicitly constraining attention between reference and generated roles. Extensive experiments demonstrate that DreamShot achieves superior scene coherence, role consistency, and generation efficiency compared to state-of-the-art text-to-image storyboard models, establishing a new direction toward controllable video model-driven visual storytelling.

\end{abstract}

\vspace{-0.2cm}
\section{Introduction}

The rapid advancement of AI-generated content (AIGC) has revolutionized visual media creation. Recent diffusion-based models have demonstrated impressive capabilities in text-to-image and text-to-video generation, producing high-fidelity single images and short clips that rival professional artistry~\cite{flux2024, wan2025wan, wu2025qwen}. However, these achievements primarily focus on static imagery or short temporal spans, while long-form, narrative-driven visual storytelling, which is the core of cinematic expression, remains largely unexplored.

Real-world visual narratives such as films and animations are structured stories composed of multiple shots, each fulfilling distinct narrative and emotional functions. Generating such multi-shot narratives requires global planning, role consistency, and scene continuity, which remain challenging for current AIGC systems~\cite{zhuang2025vistorybench, elmoghany2025survey}. Direct video synthesis is computationally expensive and redundant, generating thousands of similar frames for short moments and constraining scalability in video length, resolution, and controllability, thus hindering story-level generation.
To address these challenges, we shift the focus from dense video generation to storyboard synthesis — an efficient and controllable representation that conveys a narrative through key cinematic shots, capturing composition, perspective, and emotion without the redundancy of continuous frames.

\begin{figure}[t]
  \centering
   \includegraphics[width=1.0\linewidth]{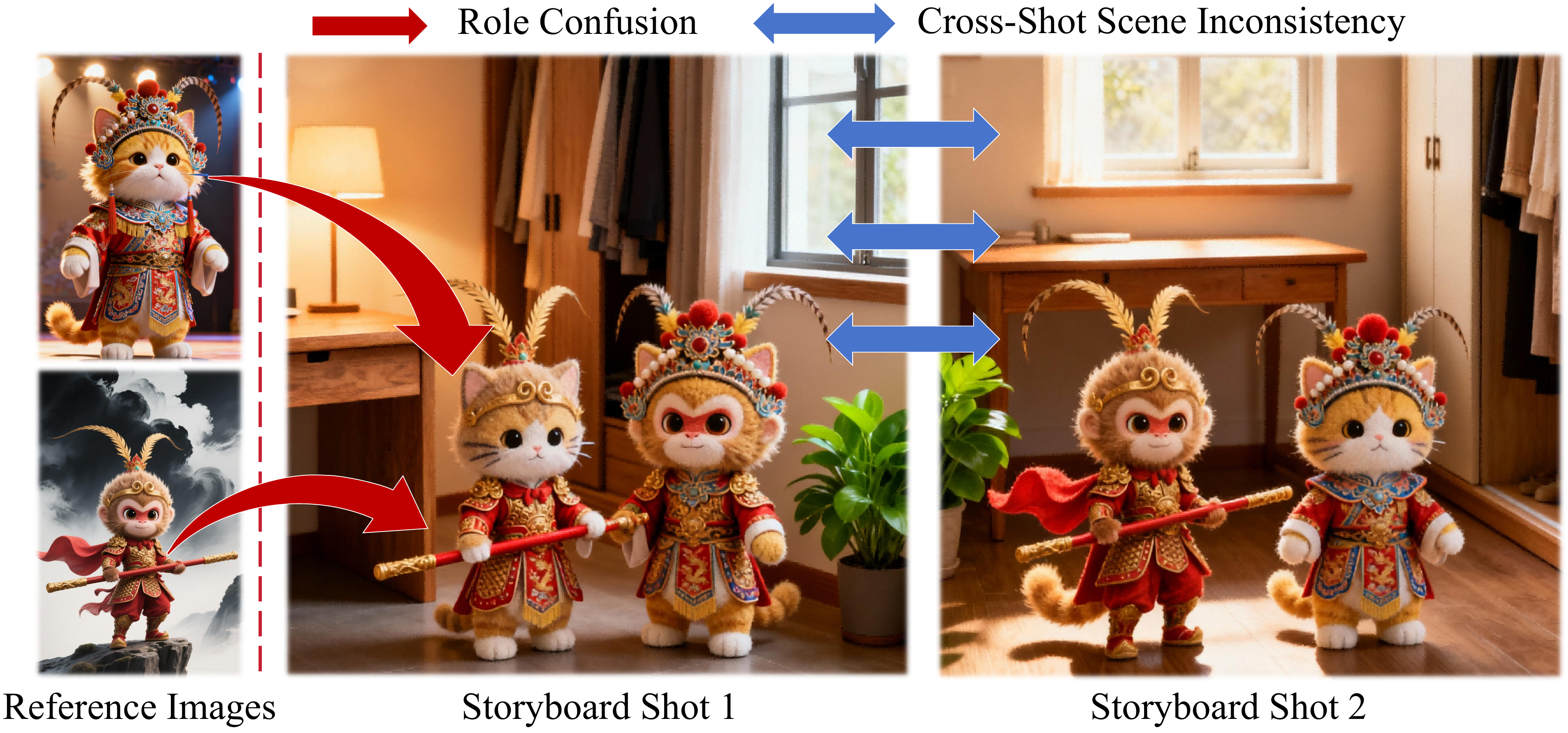}
   \vspace{-0.7cm}
   \caption{Image-based models often suffer from \textcolor{red}{role confusion} (\textcolor{red}{Red} arrrow) and \textcolor{blue}{scene inconsistency} (\textcolor{blue}{Blue} arrow) across shots.}
   \vspace{-0.5cm}
   \label{fig:intro}
\end{figure}

Early storyboard generation methods are predominantly built on image diffusion models. Works such as AnyStory~\cite{he2025anystory}, UNO~\cite{wu2025less}, InstantID~\cite{wang2024instantid}, and StoryMaker~\cite{zhou2024storymaker} focus on subject consistency via IP-Adapter~\cite{ye2023ip}, OminiControl~\cite{tan2025ominicontrol}, or ControlNet~\cite{zhang2023adding}. Although effective in preserving character identity, these models restrict controllability to portrait-level features, leaving attire, lighting, and scene layout under-constrained. The issue becomes more severe under multi-reference conditions, where feature entanglement causes role confusion, as shown in \cref{fig:intro}.
Subsequent approaches~\cite{yang2025seed, dinkevich2025story2board, rahman2023make} attempt to generate sequential storyboards from text or an initial keyframe, improving stylistic coherence but lacking fine-grained personalization and multi-identity consistency. Recent methods such as StoryDiffusion~\cite{zhou2024storydiffusion} and Story-Adapter~\cite{mao2024story} refine cross-frame attention to enhance continuity, yet image diffusion models inherently prioritize diversity over temporal stability, limiting long-range coherence.
A natural extension is to incorporate video priors. Video-based models~\cite{meng2025holocine, xiao2025captain, longcat} capture temporal dynamics and inter-frame dependencies, yielding improved motion and scene continuity. However, their dense frame synthesis introduces high computational overhead, constraining duration, resolution, and editability.
These observations lead to a key dilemma: image-based models offer flexibility but lack coherence, while video-based models offer consistency but lack efficiency. 

Building on the above insights, we propose DreamShot, a video diffusion prior based storyboard generation framework for coherent, personalized, and controllable visual storytelling. The core idea is to leverage the spatio-temporal priors of video generative models to capture temporal consistency and contextual awareness across shots while maintaining the efficiency and controllability of image-level generation. This enables DreamShot to produce globally coherent storyboards with consistent characters, scenes, and cinematic framing across multi-shot narratives.
DreamShot unifies three generation modes:  Text-to-Shot, Reference-to-Shot and Shot-to-Shot, supporting both story creation (from text or reference images) and continuation (from preceding shots), seamlessly linking local synthesis with global narrative flow.
To maintain role consistency in multi-character scenarios, DreamShot introduces a Role-Attention Consistency Loss (RACL) that explicitly aligns attention between reference images and generated shots, enforcing one-to-one role correspondence and cross-shot continuity. This effectively mitigates feature entanglement and preserves identity stability throughout the storyboard.
Complementing the model, we construct a high-quality storyboard dataset with temporally coherent shot sequences extracted from real and synthetic videos. Each sequence is paired with representative reference frames and rich shot-level annotations, supporting the supervised learning of context-aware, identity-preserving, and narratively consistent storyboards.
Together, DreamShot bridges image-based controllability with video-level coherence, offering a practical and scalable path toward long-form, cinematic visual storytelling.
In summary, our key contributions are fourfold:

\begin{itemize}
    \item We propose DreamShot, a unified video-based framework supporting multiple personalized storyboard generation modes with strong character and scene consistency
    \item We design a Role-Attention Consistency Loss to align cross-role attention, mitigating character confusion and enhancing identity consistency.
    \item We build a high-quality storyboard dataset with paired references and shot-level annotations, providing a strong foundation for personalized storyboard generation.
    \item Our method achieves state-of-the-art performance and generalizes well to out-of-distribution scenarios.
\end{itemize}

\begin{figure*}[t]
  \centering
   \includegraphics[width=1\linewidth]{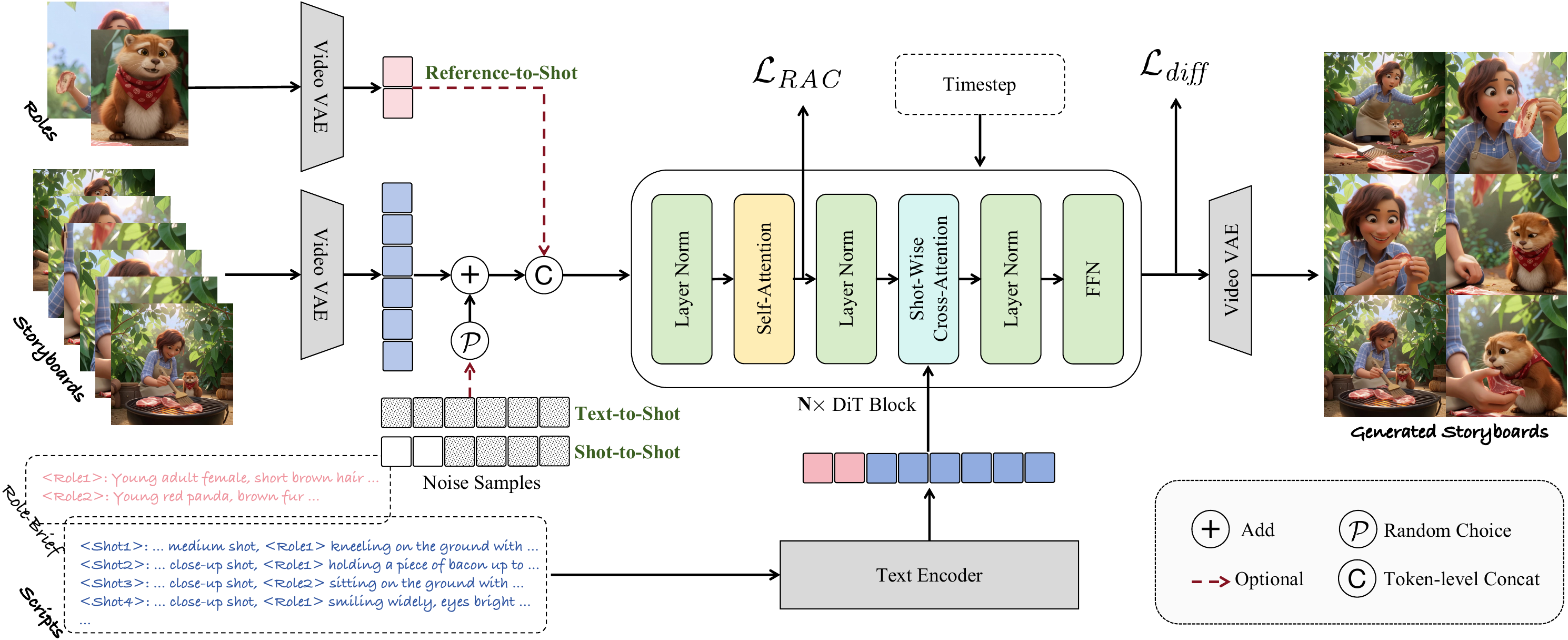}
   \vspace{-0.5cm}
   \caption{The overall framework of DreamShot consists of a Video-VAE (e.g., Wan-VAE) and a DiT module. Our model supports three generation modes: Reference-to-Shot, Text-to-Shot, and Shot-to-Shot. During inference, both the character reference image and the storyboard reference shots are introduced in a clear, noise-free form. Each shot interacts with its corresponding textual script through shot-wise cross-attention, enabling the generation of diverse storyboard shots that maintain consistent character identity and scene coherence.}
   \label{fig:method}
   \vspace{-0.2cm}
\end{figure*}

\section{Related Work}

\noindent\textbf{Storyboard Visualization.} Storyboard visualization aims to generate coherent and visually consistent shot sequences for comics or cinematic planning. Most existing approaches are built upon image-based diffusion models~\cite{rombach2022high, podell2023sdxl, flux2024}. Methods like StoryDiffusion~\cite{zhou2024storydiffusion}, Story-Adapter~\cite{mao2024story}, and Storyboard~\cite{dinkevich2025story2board} adopt training-free strategies, leveraging attention consistency and iterative refinement to generate coherent multi-shot scenes. StoryWeaver~\cite{zhang2025storyweaver} achieves customized story visualization through a Character Graph, enabling identity-consistent generation with text semantics. AnyStory~\cite{he2025anystory} and StoryMaker~\cite{zhou2024storymaker} further incorporate strong image encoders to extract facial identities and cropped character images for more personalized control. 
However, since these methods rely on image diffusion, they are constrained by the inherent priors of image models and struggle to maintain consistency across multiple shots. StoryAnchors~\cite{wang2025storyanchors} adopts a video-based paradigm for contextual coherence but supports only text or preceding shots as conditions, limiting its ability for personalized generation.

\noindent\textbf{Personalized Generation.} Personalized generation aims to produce customized and consistent outputs conditioned on one or multiple references. Some methods~\cite{han2023svdiff, gal2022image, ruiz2023dreambooth} adapt pre-trained models to embed novel concepts from a few exemplar images, while others~\cite{chen2025xverse, zhong2025mod, garibi2025tokenverse} transform reference images into token-specific modulation offsets for fine-grained personalization. In addition, several approaches~\cite{mou2025dreamo, she2025mosaic, mao2025ace++, wu2025omnigen2, huang2025dreamfuse, huang2025dreamlayer} concatenate reference features as conditional tokens into DiT-based architectures to enable direct attention interaction.
Although these methods produce high-quality personalized images, they struggle to maintain role and scene consistency across shots and provide limited control over perspective and framing.

\noindent\textbf{Video Diffusion Model.}
Compared with image models, video models are better at capturing temporal dependencies and cross-frame contextual semantics, enabling more consistent and coherent representations of dynamic scenes. Most existing video generation models~\cite{ma2025step, wan2025wan, kong2024hunyuanvideo, yang2024cogvideox} are built within diffusion-based frameworks, evolving from traditional U-Net~\cite{ronneberger2015u} architectures to transformer-based designs such as DiT~\cite{peebles2023scalable} and MMDiT~\cite{kong2024hunyuanvideo}, which greatly enhance scalability and generation quality. These models can synthesize high-quality, temporally continuous videos, but they are primarily designed for single-shot scenarios and lack explicit mechanisms to construct coherent narratives across discrete shots.
Recent studies~\cite{longcat, meng2025holocine, xiao2025captain, xiao2025videoauteur} explore multi-shot long video generation, yet these approaches typically demand heavy computational resources and can only produce a limited number of shots, making them less suitable for controllable or editable storyboard creation.
\section{Methodology}

As illustrated in \cref{fig:method}, DreamShot is built upon a video diffusion backbone, consisting of a spatio-temporal video VAE and a Diffusion Transformer (DiT).
Each storyboard group consists of the following elements: $K$ reference roles $\{I_{ref}^{(k)}\}_{k=1}^{K}$ appearing in the storyboards, each with a prompt $C_{ref}^k$; and $S$ storyboard shots $\{I_{shot}^{(s)}\}_{s=1}^{S}$, each with an annotation $C_{shot}^s$.  As shown in \cref{fig:method}, shots are denoted as `\textit{\textless Shot x\textgreater}', and reference roles as `\textit{\textless Role x\textgreater}'.

\subsection{DreamShot Framework}
\label{dreamshot framework}
Modern video VAEs (e.g., Wan-VAE~\cite{wan2025wan}) perform temporal compression by encoding every $T$ consecutive frames into a single latent representation (commonly $T$=4), substantially reducing computation while preserving temporal structure. Importantly, video VAEs adopt causal spatiotemporal encoding, such property inherently models a forward-evolving temporal process.
To align storyboard generation with both the temporal compression and the causal temporal structure of the video VAE, we repeat each storyboard shot $T$ times (except the first shot) before encoding. This produces a temporally ordered latent sequence, where each shot acts as a stable temporal segment consistent with the VAE’s causal encoding behavior: $z_{shot} \in \mathbb{R}^{s\times d\times h\times w}$, where $s$ is the number of shots and $d$, $h$, $w$ are the latent dimensions.
This converts the originally independent storyboard images into a coherent latent stream consistent with the VAE’s causal structure, allowing the DiT to interpret the storyboard as a coherent temporal narrative rather than isolated frames.
Each reference image is encoded independently using the same video VAE. We treat these reference latents as the preceding temporal anchor for the storyboard sequence, denoted as $z_{ref} \in \mathbb{R}^{k\times d\times h\times w}$, where k is the number of reference identities. In parallel, the text descriptions corresponding to each reference image and storyboard shot are encoded with umT5 text encoder~\cite{chung2023unimax}, providing the condition embeddings, $C_t$.
This structure produces temporally aligned latent tokens and text embeddings, enabling DreamShot to fully exploit the spatiotemporal priors in the video diffusion backbone.

To fully exploit the 3D rotary position encoding (RoPE)~\cite{su2024roformer} employed in video diffusion models, we concatenate the reference-image tokens before the shot tokens, forming $z_t=[z_{ref}, z_{shot}]$. Unlike image-based diffusion models that rely on 2D RoPE, video models encode spatial and temporal positions, enabling them to model long-range temporal evolution. By arranging reference tokens first and storyboard-shot tokens afterward, we explicitly map the input sequence into a temporal ordering: \emph{references occur first, and followed by the story progression across shots}. This simple but crucial design enables the DiT to propagate role identity forward in time, maintaining character consistency and logical narrative evolution capabilities, which are fundamentally absent in image-based models.

\noindent\textbf{Shot-Wise Cross-Attention.}
Within the $N$ DiT blocks, self-attention is computed jointly over reference and shot tokens, allowing the model to inject reference-image identity cues directly into storyboard representations:
\begin{equation}
    z_s=Softmax(\frac{Q_sK_s^T}{\sqrt{d}})V_s,
\end{equation}
where $Q_s, K_s, V_s$ are obtained through linear projections of the token $z_t$. The resulting features are passed to a shot-wise cross-attention layer, where each storyboard shot independently attends to its corresponding text embedding for fine-grained vision–language alignment:
\begin{equation}
    z_c^i=Softmax(\frac{Q_c^i(K_c^i)^T}{\sqrt{d}})V_c^i, i=0,\cdots,k+s,
\end{equation}
where $Q_c$ is linearly projected from $z_s$ and $K_c, V_c$ are linearly projected from $C_t$.

\subsection{Learning Strategy}
Based on the architecture in \cref{dreamshot framework}, we perform mixed-mode training across all generation modes, i.e., Text-to-Shot, Reference-to-Shot and Shot-to-Shot.
In the reference-to-shot mode, noise is applied only to the storyboard shot tokens at timestep $t$:
\begin{equation}
    z_t= (1-t)z_{shot} + tz_n, z_n \sim \mathcal{N} (0,1),
\end{equation}
where $z_n$ denotes a noise sample.

With the timestep $t$, reference tokens $z_{ref}$ and text prompts $C_t$ as conditions,
the diffusion model trains a network $\epsilon_\theta$ to regress the velocity field $\epsilon_\theta(z_t, C_t, t)$ by minimizing the Flow Matching objective:
\begin{equation}
    \mathcal{L}_\textit{diff}=\mathbb{E}_{t, (z_t, C_t)\sim D, z_n\sim \mathcal{N} (0,1)}[||\epsilon-\epsilon_\theta(z_t, C_t, t)||],
\end{equation}
where the target velocity field is $\epsilon=z_n-z_{shot}$.

Similarly, in the text-to-shot mode, each storyboard shot is perturbed with noise at the same timestep, where the model predicts the noise at each timestep under textual guidance. In contrast, in the shot-to-shot mode, the preceding shots are used as reference inputs without any noise injection, representing a clear and lossless condition to guide the generation of subsequent shots.

\noindent\textbf{Role-Attention Consistency Loss.} 
In cinematic storyboards, multiple reference roles often appear simultaneously. However, existing models mainly focus on identity preservation and tend to suffer from role confusion when handling multiple identities, where features of different roles such as faces and clothing are incorrectly merged into a single role. To address this, we propose a Role-Attention Consistency Loss (RACL), which enforces similarity constraints among corresponding role representations during training, thereby reducing the likelihood of cross-role feature confusion.

\begin{figure}[t]
  \centering
   \includegraphics[width=0.92\linewidth]{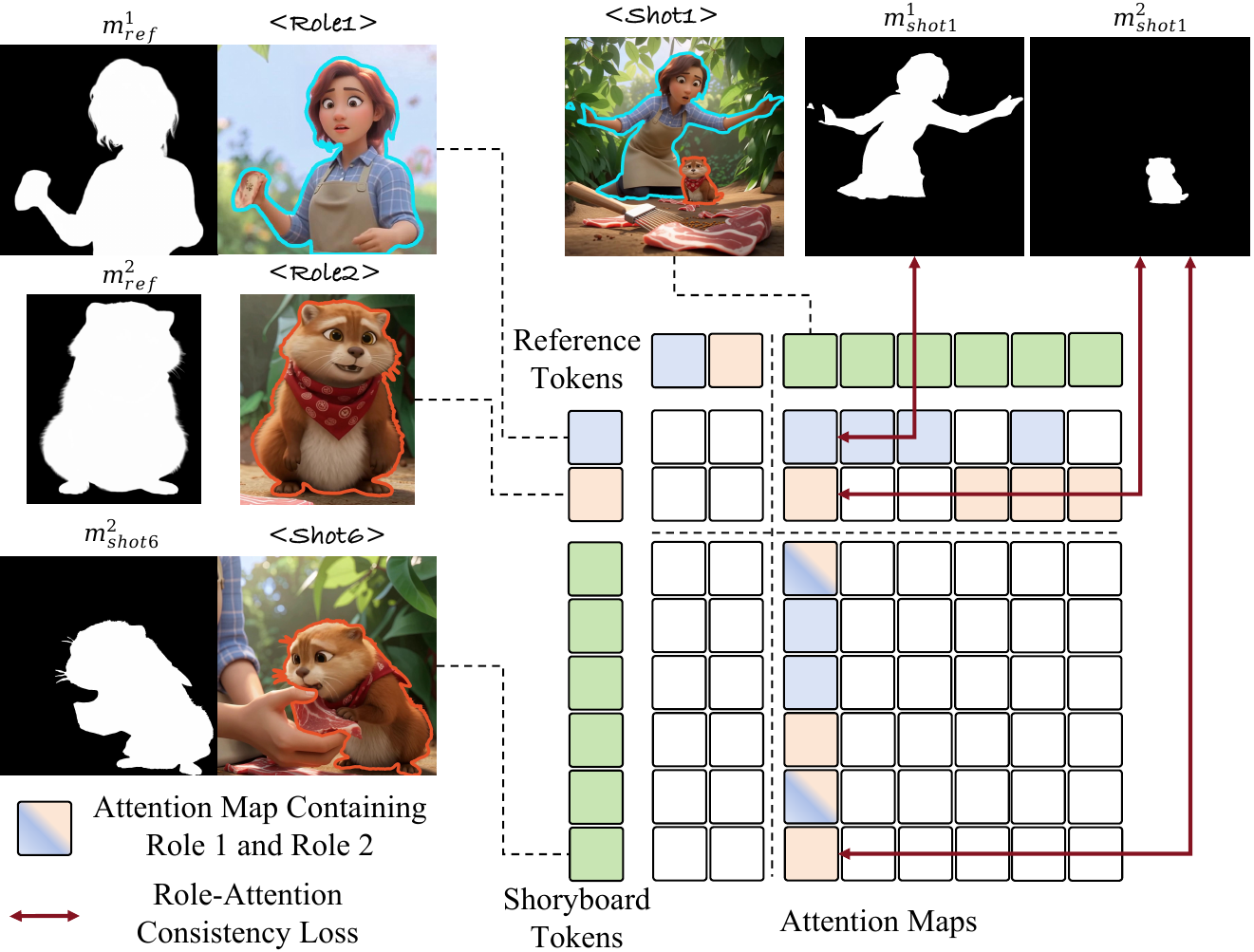}
   \vspace{-0.2cm}

   \caption{Illustration of the Role-Attention Consistency Loss (RACL). RACL supervises the attention maps between each reference image and each storyboard shot, ensuring that each role attends to its corresponding spatial regions while suppressing entanglement across different identities. By enforcing one-to-one role correspondence and penalizing mixed-role attention, RACL improves role grounding, enhances cross-shot identity consistency, and mitigates multi-reference confusion.
   }
   \label{fig:rac}
   \vspace{-0.2cm}
\end{figure}

We obtain role masks for both reference and storyboard images to distinguish multiple roles, as shown in \cref{fig:rac}. Reference masks are generated via saliency-based extraction, and storyboard roles are segmented using grounding-based segmentation~\cite{ren2024grounded}. We then employ ArcFace~\cite{deng2019arcface} and a VLM~\cite{guo2025seed1} to establish one-to-one correspondences between reference and storyboard roles, yielding paired masks $(m_{ref}^k, m_{s}^k)$ for each role $k$ in shot $s$.

In the DiT self-attention, each role in the reference image attends to its corresponding counterpart in the storyboard shot. The correlation is represented by the resulting attention map, as shown in \cref{fig:rac}. For a reference role $k$, the attention between this role $r_k$ and its counterpart in storyboard shot $s_k$ across $N$ DiT blocks can be defined as:
\begin{equation}
    A_{r_k-s_k}=\frac{1}{N}\sum_{n=1}^{N}\sum_{i=1}^{m_{ref}^k} Softmax(Q_iK_{s_k}^T/\sqrt{d}),
\end{equation}
which represents the role attention between the reference and storyboard images. We expect the attention to focus on role $k$ itself rather than being dispersed to other regions. Therefore, we apply the mask $m_s^k$ as a supervision constraint to enforce this focus.
Similarly, the attention between role $k$ in shot $s$ ($s_k$) and the same role in shot $s'$ ($s'_k$) can also be computed as
\begin{equation}
    A_{s'_k-s_k}=\frac{1}{N}\sum_{n=1}^{N}\sum_{i=1}^{m_{s'}^k} Softmax(Q_iK_{s_k'}^T/\sqrt{d}).
\end{equation}
We apply the same supervision constraint to prevent roles from other shots from being confused with those in shot $s$.
Accordingly, the final loss can be formulated as:
\begin{equation}
\begin{aligned}
    \mathcal{L}_{RAC}=\frac{1}{S}\sum_{s=1}^S\frac{1}{K}\sum_{k=1}^K \left( \operatorname{BCE}(A_{r_k-s_s}, m_s^k) \right.\\
    + \left.\operatorname{BCE}(A_{s'_k-s_k}, m_s^k)\right),
\end{aligned}
\end{equation}
where $\operatorname{BCE}$ denotes the binary cross entropy loss.
\begin{equation}
    \operatorname{BCE}(A, m) = -[m\log A+(1-m)\log (1-A)].
\end{equation}
By interating $\mathcal{L}_{RAC}$ into the training objective, our model is encouraged to learn precise role correspondences between reference and storyboard shots. This significantly mitigates role confusion and enhances role consistency across generated shots. The final training objective 
is defined as $\mathcal{L} = \lambda \mathcal{L}_{RAC} + \mathcal{L}_{diff},$
where $\lambda$ is the weigh term.

\begin{figure*}[t]
  \centering
   \includegraphics[width=0.88\linewidth]{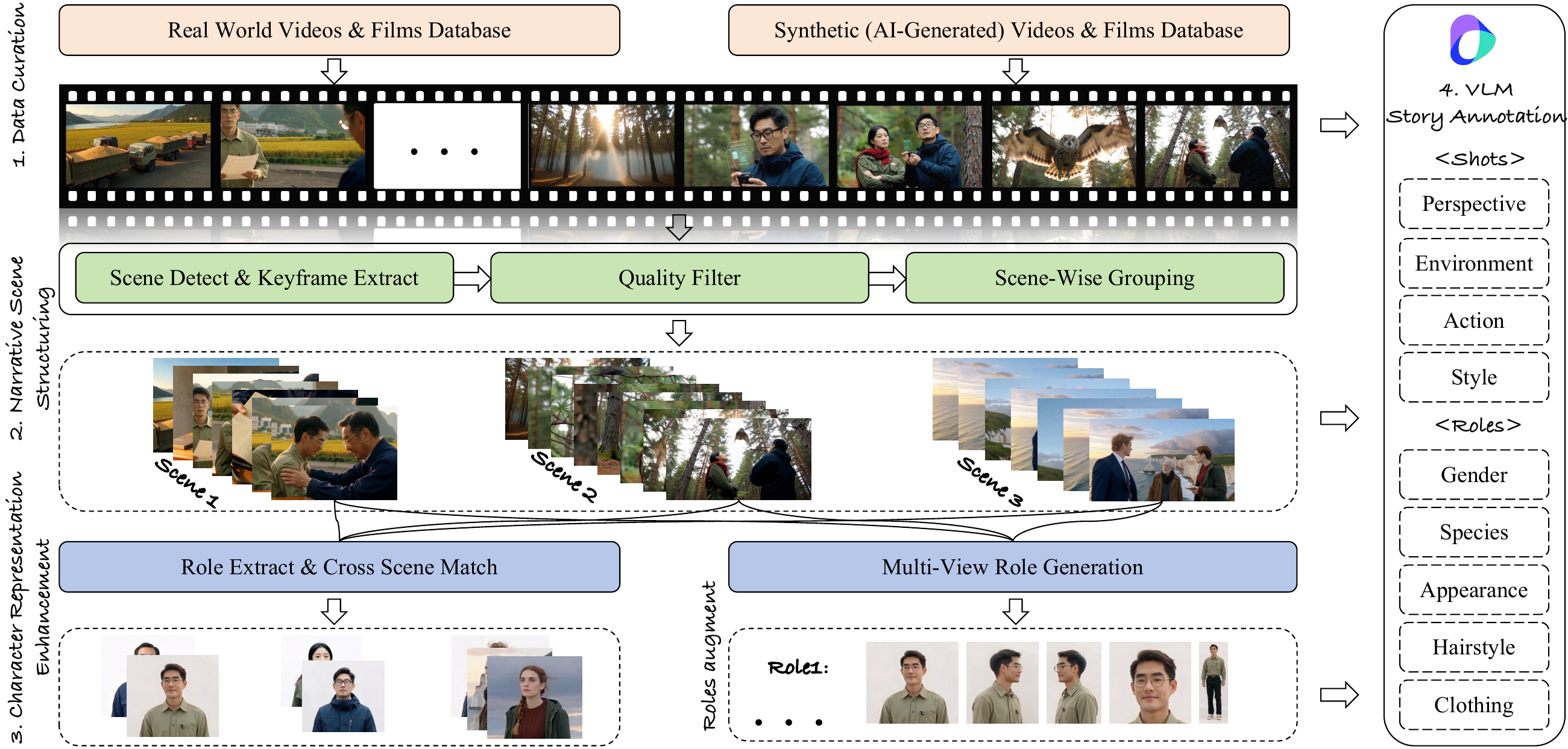}
    \vspace{-0.2cm}
   \caption{Overview of the data construction pipeline. The process consists of four main stages: data source selection, keyframe extraction and scene-wise grouping, role extraction and augmentation, and story annotation. Rigorous quality control is applied at each stage to ensure the consistency and reliability of the dataset.}
   \label{fig:data_pipe}
\end{figure*}

\noindent\textbf{Reference-Free CFG.} 
Similar to classifier-free guidance~\cite{ho2022classifier} (CFG), we extend it to a reference-free guidance scheme. During training, we randomly drop text or image conditions to create negative samples. At inference, outputs generated with negative prompts and without reference images are used to guide the final storyboard generation.
The denoised latent at timestep $t$ is defined as
\begin{equation}
\begin{aligned}
    z_{t-1} = z^{\varnothing, neg}_{t} + \omega_1(z^{ref, neg}_{t}-z^{\varnothing, neg}_{t}) \\
    +\omega_2(z^{ref,pos}_{t}-z^{ref, neg}_{t}),
\end{aligned}  
\end{equation}
where $z^{\varnothing, neg}_{t}$ denotes the unconditional denoising output with negative prompts only,  $z^{ref, neg}_{t}$ represents the output conditioned on reference images and negative prompts and $z^{ref,pos}_{t}$ corresponds to the output conditioned on both reference images and positive prompts. The coefficients $\omega_1$ and $\omega_2$ are  weight terms.
This mechanism enhances the consistency between the roles in the generated shots and their corresponding reference roles.

\section{Storyboard Dataset}
As shown in \cref{fig:data_pipe}, we develop a scalable storyboard-generation pipeline, consisting of four stages: data curation, narrative scene structuring, character representation enhancement, and story annotation.
Further details are provided in the supplementary materials.

\noindent\textbf{\textit{Data Curation.}} 
Storyboard data are scarce and need to be extracted from cinematic videos. Existing open-source datasets~\cite{wang2025koala,nan2024openvid} rarely contain storyboard-style structures, while high-quality narrative videos are difficult to access. To mitigate this, we collected 40K high-quality videos from web and supplemented them with 50K AIGC-generated videos created by advanced video tools~\cite{gao2025seedance} and diverse prompts. These synthetic videos are easier to obtain and cover a wide range of styles, scenes, and narratives, providing rich material for storyboard generation.


\noindent\textbf{\textit{Narrative Scene Structuring.}} 
Next, unlike method~\cite{wang2025storyanchors} that uniformly sample frames as storyboards, we extract representative, narratively coherent, and scene-consistent shots from raw videos. Specifically, we use PySceneDetect~\cite{PySceneDetect} to segment videos based on shot transitions and extract representative keyframes from each shot. We apply the Laplacian operator and quality assessment~\cite{aes} to remove low-quality and contentless frames, and use a VLM~\cite{guo2025seed1} to cluster narratively coherent frames within the same scene into storyboard groups. Each long video is ultimately divided into multiple storyboard groups based on scene boundaries.


\begin{table}[t]
    \vspace{-0.4cm}
    \centering
    \footnotesize
    \renewcommand{\arraystretch}{0.95}
    \setlength{\tabcolsep}{0.5mm}{
    \begin{tabular}{lcccccccc}
    \toprule
        \multirow{2}*{Methods} & \multicolumn{2}{c}{CIDS(Character)} && \multicolumn{2}{c}{CSD(Style)} & AES & Alignment  \\
        \cline{2-3} \cline{5-6}
        \rule{0pt}{3ex} ~ & Self$\uparrow$ & Cross$\uparrow$  & & Self$\uparrow$  & Cross$\uparrow$  & Score$\uparrow$  & Score$\uparrow$  \\
    \midrule 
        \rowcolor{palered} 
        \multicolumn{8}{c}{Reference-to-Shot} \\
        Story-Adapter~\cite{mao2024story} & 36.1 & 25.5 && 51.9 & 36.7 & 5.44 & 1.84 \\
        DreamO~\cite{mou2025dreamo} & 47.2 & 40.0 & & 58.3 & 39.5 & 5.45 & 3.29 \\
        UNO~\cite{wu2025less} & 38.6 & 28.6 && 56.2 & 40.3 & 5.28 & 3.27 \\
         \rowcolor{mygray}
        DreamShot & \textbf{48.7} & \textbf{51.6} && \textbf{64.5} & \textbf{45.1} & \textbf{5.50} & \textbf{3.39}\\
    \midrule
        \rowcolor{palegreen} 
        \multicolumn{8}{c}{Text-to-Shot} \\
        Story-Adapter~\cite{mao2024story} & 45.4 & - & & 44.2 & - & 5.41 & 1.98 \\
        StoryDiffusion~\cite{zhou2024storydiffusion} & 45.4 & - && 56.9 & - & 5.38 & 1.82 \\
        Story2Board~\cite{dinkevich2025story2board} & 44.4 & - & & 62.6 & - & 5.27 & 2.95 \\
         \rowcolor{mygray}
        DreamShot & \textbf{46.8} & - && \textbf{65.4} & - & \textbf{5.46} & \textbf{3.22} \\
    \midrule
        \rowcolor{paleblue} 
        \multicolumn{8}{c}{Shot-to-Shot} \\
        Story-Adapter~\cite{mao2024story} & \textbf{42.2} & - & & 52.2 & - & 5.45 & 2.27 \\
        UNO~\cite{wu2025less} & 36.7 & - & & 55.8 & - & 5.12 & 3.28 \\
         \rowcolor{mygray}
        DreamShot & 40.4 & - & & \textbf{62.3} & - & \textbf{5.46} & \textbf{3.31} \\
    \bottomrule
    \end{tabular}
    }
    \vspace{-0.2cm}
    \caption{Quantitative evaluation on DreamShot test set, including three modes: Reference-to-Shot, Text-to-Shot, and Shot-to-Shot}
    \label{tab:sota-dreamshot}
    \vspace{-0.4cm}
\end{table}

\noindent\textbf{\textit{Character Representation Enhancement.}} 
We extract reference roles from each storyboard group, identifying and aggregating main characters by portrait ID.
To reduce the copy-and-paste artifacts that may occur during reference-to-shot generation, we introduce a cross-storyboard matching strategy: using ArcFace~\cite{deng2019arcface}, we retrieve the same characters from other storyboard groups to serve as reference roles for the current group.
Furthermore, to enhance the diversity of the reference roles, we employ video-based augmentation~\cite{gao2025seedance} or reference-image generation~\cite{batifol2025flux, seedream2025seedream}, expanding each roles into multiple reference forms like half-body, full-body, close-up, and side-view portraits.

\noindent\textbf{\textit{Story Annotation.}} Based on the storyboards, we employ a VLM~\cite{guo2025seed1} to annotate each shot in terms of perspective, environment, action, and style, ensuring narrative coherence throughout the sequence. Each reference image is also annotated with attributes like gender, appearance, and clothing, resulting in a comprehensive storyboard dataset.

\section{Experiments}
\subsection{Implementation Details}


\begin{table}[t]
    \vspace{-0.4cm}
    \centering
    \footnotesize
    \renewcommand{\arraystretch}{0.95}
    \setlength{\tabcolsep}{0.5mm}{
    \begin{tabular}{lcccccccc}
    \toprule
        \multirow{2}*{Methods} & \multicolumn{2}{c}{CIDS(Character)} && \multicolumn{2}{c}{CSD(Style)} & AES & Alignment  \\
        \cline{2-3} \cline{5-6}
        \rule{0pt}{3ex} ~ & Self$\uparrow$ & Cross$\uparrow$  & & Self$\uparrow$  & Cross$\uparrow$  & Score$\uparrow$  & Score$\uparrow$  \\
    \midrule 
        \rowcolor{palered} 
        \multicolumn{8}{c}{Reference-to-Shot} \\
        Story-Adapter~\cite{mao2024story} & 60.5 & 46.0 && 54.8 & \textbf{45.6} & 4.99 & 2.69 \\
        SEED-Story~\cite{yang2025seed} & 58.7 & 28.7 && \textbf{74.8} & 22.7 & 3.82 & 1.23  \\
        UNO~\cite{wu2025less} & 62.0 & 48.5 && 60.2 & 39.1 & 5.23 & 3.01 \\
        OmniGen2~\cite{wu2025omnigen2} & \underline{64.7} & \underline{54.8} && 60.0 & \underline{45.4} & 5.25 & 3.10 \\
        QwenImage~\cite{wu2025qwen} & 57.4 & 47.5 && 59.3 & 38.1 & \textbf{5.50} & \underline{3.28} \\
         \rowcolor{mygray}
        DreamShot & \textbf{65.6} & \textbf{60.2} && \underline{61.8} & 43.9 &  \underline{5.40} & \textbf{3.37} \\
    \bottomrule
    \end{tabular}
    }
    \vspace{-0.2cm}
    \caption{Latest quantitative evaluation on VistoryBench.}
    \label{tab:sota-vistory}
    \vspace{-0.4cm}
\end{table}

\noindent\textbf{Hyperparameters.}
We adopt Wan2.1-14B~\cite{wan2025wan} as the base model. During training, the first 10k iterations are performed at 192p resolution with the Role-Attention Loss ($\lambda=0.2$) to guide the model in learning character-role correspondence. In the following 10k iterations, the resolution is increased to 480p, and the model is trained only with \(\mathcal{L}_{diff}\) using a learning rate of \(3\times10^{-5}\), the AdamW~\cite{loshchilov2017decoupled} optimizer, and a LoRA~\cite{hu2022lora} rank of 256. During inference, \(\omega_1\) and \(\omega_2\) are set to 4 and 5, respectively. The entire training process is conducted on 8 A800 GPUs and takes approximately 48 hours.

\begin{figure*}[t]
  \centering
   \includegraphics[width=0.87\linewidth]{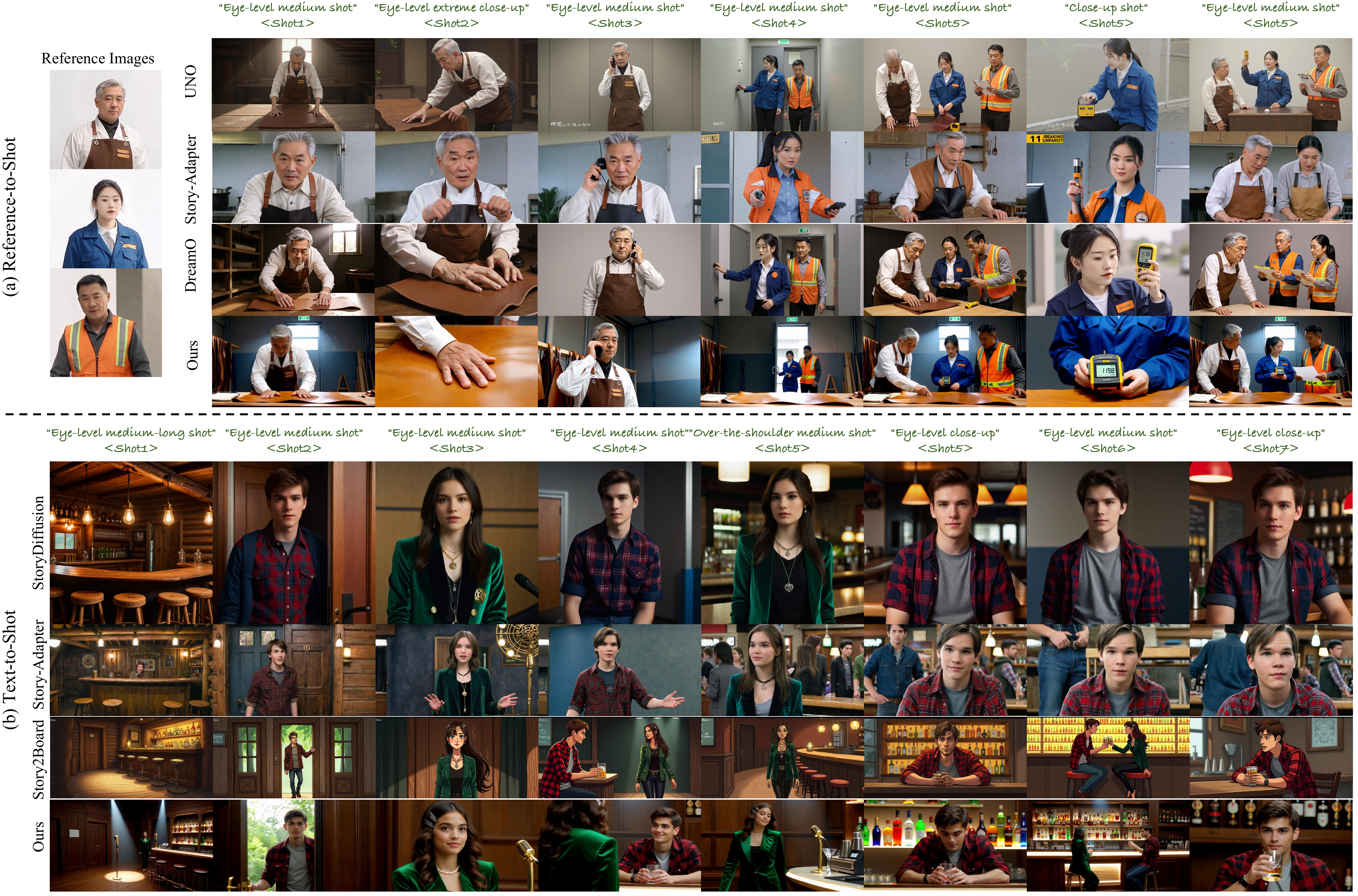}
   \vspace{-0.3cm}
   \caption{Qualitative comparisons with existing methods. (a) Reference-to-Shot generation. Compared with other methods, our approach adapts well to shot transitions while maintaining consistent environmental layout across scenes. (b) Text-to-Shot generation. Our results reflect better cinematic storytelling rather than portrait-like compositions, while achieving superior character consistency across shots.}
   \label{fig:visual}
   \vspace{-0.3cm}
\end{figure*}

\noindent\textbf{Benchmarks.}
We comprehensively evaluate our method on the DreamShot Test Set and VistoryBench~\cite{zhuang2025vistorybench}. The DreamShot Test Set contains 100 storyboard samples, each including reference images, textual descriptions, and examples of previous frames. We further conduct evaluations on VistoryBench, which consists of 80 story samples, 1,317 storyboard shots, and 509 reference images collected from diverse sources such as film and television scripts, literary classics, world legends, novels, and picture books.

\noindent\textbf{Evaluation metrics.} Following VistoryBench~\cite{zhuang2025vistorybench},  We evaluate Cross-Similarity and Self-Similarity, measuring the resemblance between generated and reference images and the consistency among generated shots. Character Identification Similarity (CIDS) computes the average cosine similarity of character features across shots to assess identity consistency, while Contrastive Style Descriptors (CSD)~\cite{somepalli2024measuring} measure style and scene coherence using CLIP-extracted features.
We further report an Alignment Score based on GPT to evaluate the correspondence between generated shots and textual descriptions, covering scene layout, actions, and camera perspective. The AES Score~\cite{aes} is also used to assess the aesthetic quality of each shot.

\subsection{Comparisons with Existing Methods}

\noindent\textbf{Quantitative results.}
We evaluate our method on the DreamShot Test Set across three personalized generation settings: Reference-to-Shot, Text-to-Shot, and Shot-to-Shot, comparing it with state-of-the-art methods including Story-Adapter~\cite{mao2024story}, UNO~\cite{wu2025less}, DreamO~\cite{mou2025dreamo}, StoryDiffusion~\cite{zhou2024storydiffusion}, and Story2Board~\cite{dinkevich2025story2board}. As shown in \cref{tab:sota-dreamshot}, our method outperforms existing approaches across multiple metrics, achieving a notable $6.2$ improvement in the CSD-Self score over the second-best method in the reference-to-shot setting. This demonstrates its superior scene and style consistency across storyboard shots, producing results more aligned with visual storytelling. 
Furthermore, we evaluate our method on VistoryBench~\cite{zhuang2025vistorybench}, as shown in \cref{tab:sota-vistory}. On this multi-scene and multi-style benchmark, our approach also achieves superior performance, surpassing the second-best method by $8.2$ in the CIDS-Cross score, demonstrating stronger preserves character identity and maintains stronger 

\noindent\textbf{Qualitative results. }
\cref{fig:visual} shows qualitative comparisons between our method and others. (a) In the Reference-to-Shot mode, our approach effectively preserves the reference character identity while maintaining scene consistency across shot transitions. (b) In the Text-to-Shot mode, our results exhibit stronger cinematic quality and maintain consistent character appearance across different shots.

\subsection{Ablation Study}

\begin{table}[t]
    \centering
    \footnotesize
    \setlength{\tabcolsep}{0.5mm}{
    \begin{tabular}{lcccccccc}
    \toprule
        \multirow{2}*{DreamShot} & \multicolumn{2}{c}{CIDS(Character)} && \multicolumn{2}{c}{CSD(Style)} & AES & Alignment  \\
        \cline{2-3} \cline{5-6}
        \rule{0pt}{3ex} ~ & Self$\uparrow$ & Cross$\uparrow$  & & Self$\uparrow$  & Cross$\uparrow$  & Score$\uparrow$  & Score$\uparrow$  \\
    \midrule 
        \rowcolor{palered} 
        w/o $\mathcal{L}_{RAC}$ & \multicolumn{7}{c}{Reference-to-Shot} \\
        Phase-Pos & 43.6 & 20.0 & & \textbf{67.2} & 36.2 & 5.47 & 2.96 \\
        Neg-Pos & 47.2 & 40.9 & & 60.5 & 35.0 & 5.20 & 3.03 \\
        Last-Pos & 43.5 & 43.5 & & 65.3 & 43.4 & \textbf{5.53} & 3.26 \\
        \midrule
        Ours w/o $\mathcal{L}_{RAC}$ & 46.7 & 48.3 & & 64.2 & 43.7 & 5.49 & 3.26  \\ 
         \rowcolor{mygray}
        Ours & \textbf{48.7} & \textbf{51.6} && 64.5 & \textbf{45.1} & 5.50 & \textbf{3.39}\\
    \bottomrule
    \end{tabular}
    }
  \vspace{-0.2cm}
    \caption{Qualitative analysis of different positional incorporation strategies for reference images (upper part) and the effectiveness of $\mathcal{L}_{RAC}$ (lower part).}
    \label{tab:ablation}
\end{table}

\begin{figure}[t]
  \centering
  \vspace{-0.2cm}
   \includegraphics[width=0.85\linewidth]{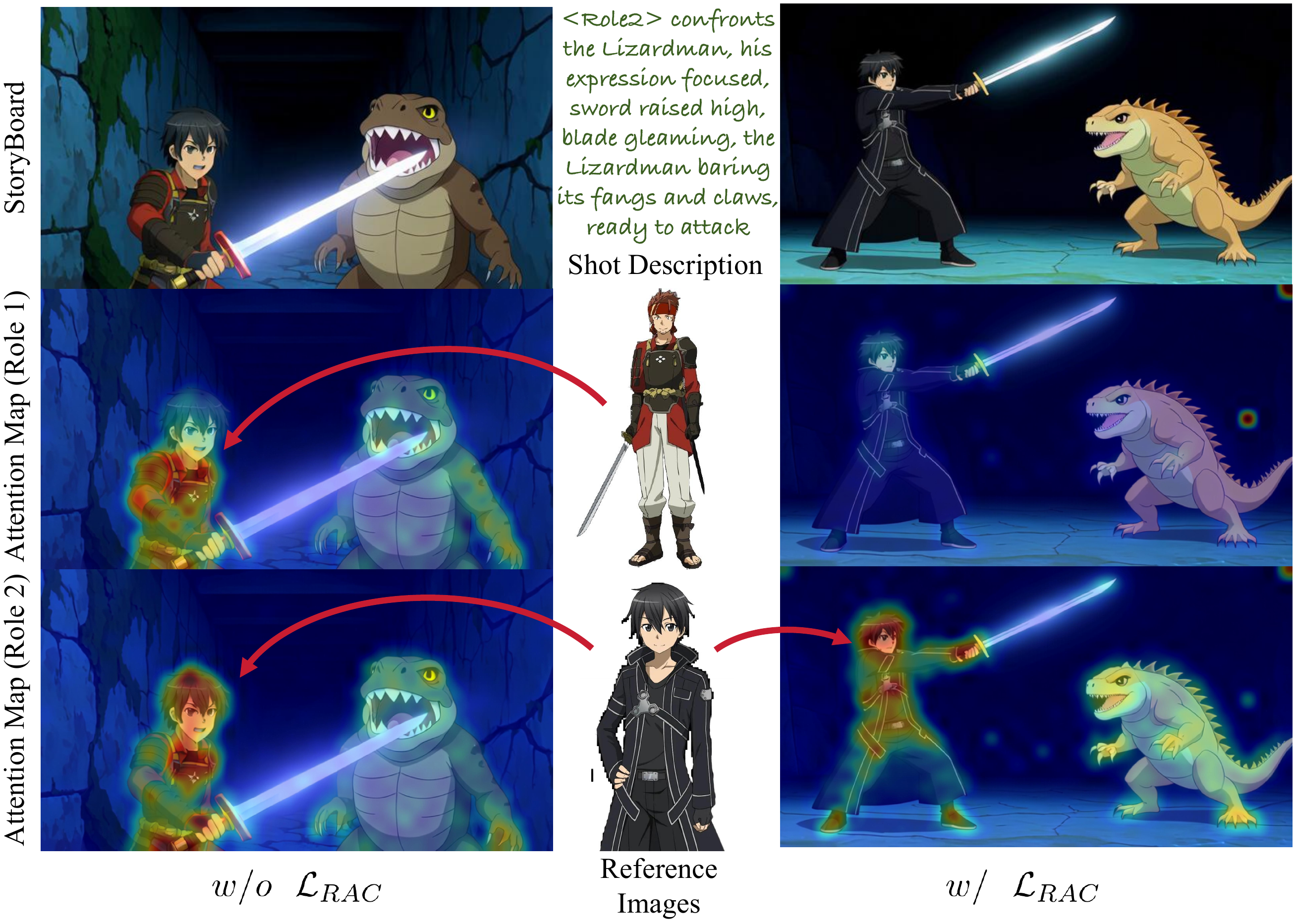}
  \vspace{-0.2cm}
   \caption{Visualization of attention maps with and without $\mathcal{L}_{RAC}$ under the two-reference images. Without $\mathcal{L}_{RAC}$, the attention of different reference roles tends to overlap, causing confusion in facial features and clothing.}
   \label{fig:attn_map}
   \vspace{-0.4cm}
\end{figure}

\noindent\textbf{Role-Attention Consistency Loss. } As shown in \ref{tab:ablation}, we compare our model with `Ours w/o $\mathcal{L}_{RAC}$. The results show that incorporating $\mathcal{L}_{RAC}$ improves role consistency and enhances alignment with reference images after role disentanglement, achieving a $3.3$ gain in the CIDS-Cross score. To further examine the effect, we visualize the attention maps during inference, as illustrated in \cref{fig:attn_map}. Without $\mathcal{L}_{RAC}$, the attention of different reference characters tends to overlap, causing confusion in facial features and clothing. In contrast, $\mathcal{L}_{RAC}$ effectively alleviates this issue, leading to clearer attention separation and higher role consistency.

\noindent\textbf{Position Embedding of Reference Images.} To better distinguish the reference image from the generated storyboard, we compare several positional embeddings applied to the reference images, as summarized in  \cref{tab:ablation}. Specifically, we test a phase offset (Phase-Pos) in the RoPE~\cite{su2024roformer} encoding of the reference image, which directly adds an angular shift to each frequency component to globally rotate the embedding phase, and a negative offset (Neg-Pos), which shifts the effective positional index backward to simulate a spatial delay along the positional axis. In addition, we also experiment with placing the reference image at the end of the generated storyboard sequence (Last-Pos) to mitigate the instability introduced by VAE encoding. 
Introducing additional offsets pushes the reference embeddings away from the storyboard space, causing a significant drop in consistency, especially in the Phase-Pos setting, where the CIDS-Cross score decreases by $23.5$. Placing the reference at the end slightly mitigates compression and improves aesthetics but reduces consistency and text alignment. Therefore, we position the reference at the beginning as a preceding shot to better guide subsequent storyboard generation.

\begin{figure}[t]
  \centering
   \includegraphics[width=0.49\linewidth]{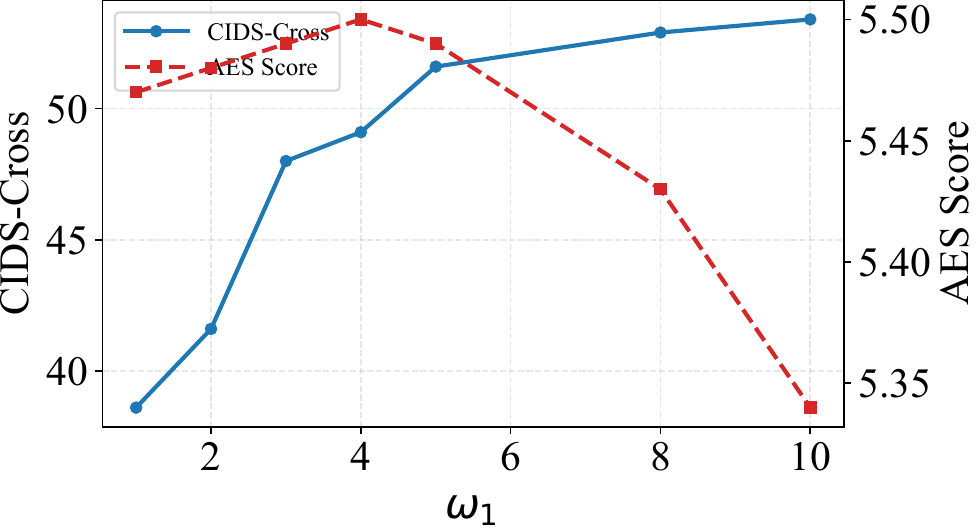}
   \includegraphics[width=0.49\linewidth]{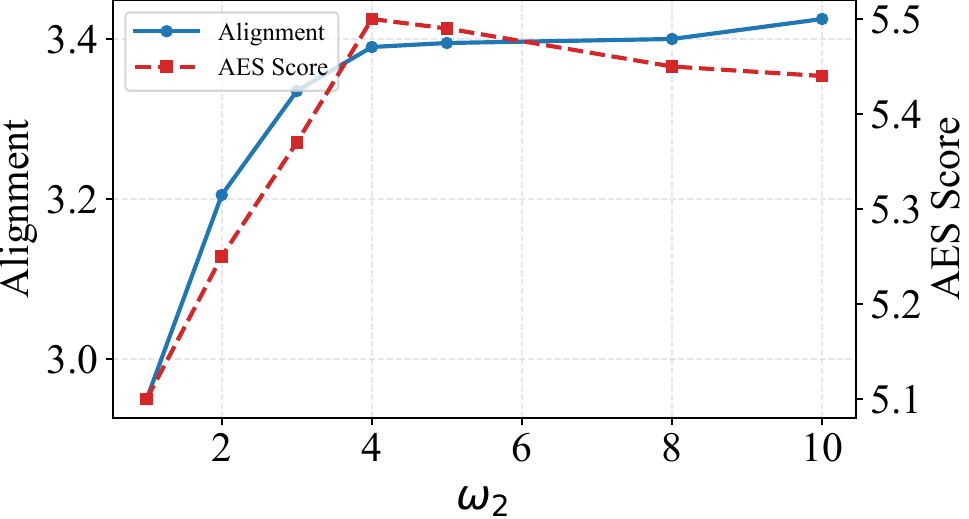}
  \vspace{-0.2cm}
   \caption{The effect of the $\omega_1$ and $\omega_2$ in reference-free CFG.}
   \label{fig:cfg_img}
\end{figure}

\noindent\textbf{Reference-Free CFG.} As shown in \cref{fig:cfg_img}, we investigate the impact of $\omega_1$ and $\omega_2$ in the reference-free CFG. $\omega_1$ plays a key role in controlling character consistency: when $\omega_1=1$, no guidance is applied, resulting in poor character consistency; increasing $\omega_1=1$ consistency but overly large values (e.g. $\omega_1=8$) significantly degrade generation quality and lower the AES score. The parameter $\omega_2$ primarily affects text alignment. Based on these observations, we adopt $\omega_1=4, \omega_2=5$ as the optimal configuration.

\begin{figure}[t]
  \centering
  \vspace{-0.2cm}
   \includegraphics[width=0.75\linewidth]{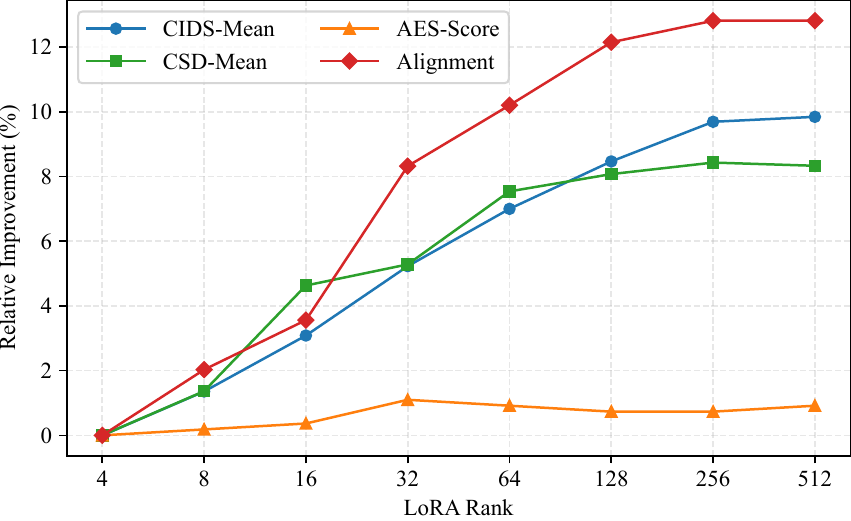}
   \vspace{-0.2cm}
   \caption{The effectiveness of LoRA Rank.}
   \label{fig:lora rank}
   \vspace{-0.4cm}
\end{figure}

\noindent\textbf{LoRA Rank.} \cref{fig:lora rank} shows the relative improvement across different LoRA ranks. The Alignment Score is most affected, with higher ranks improving text responsiveness, while the AES Score shows relatively minor variation.

\section{Conclusion}
In this paper, we propose DreamShot, a personalized storyboard generation framework that leverages the consistency priors of video models to produce more coherent and cinematic storyboards. We construct a diverse dataset of cinematic storyboard samples and support multiple generation modes, including Reference-to-Shot, Text-to-Shot, and Shot-to-Shot. Furthermore, we introduce the Role-Attention Consistency Loss to strengthen role correspondence between reference images and generated shots, effectively reducing character confusion. Experimental results demonstrate that our method outperforms existing approaches across multiple benchmarks. The current performance is limited by the base model and the length of available storyboard data, leading to suboptimal results on extremely long shot sequences.

\section{Acknowledgment}
This work is supported in part by the National Key
R\&D Program of China (NO.~2024YFB3908503 and 2024YFB3908500), in part by the National Natural Science Foundation of China (No.~62322608), and in part by the Major Key Project of PCL (Grant No.~PCL2025A17). 
{
    \small
    \bibliographystyle{ieeenat_fullname}
    \bibliography{main}
}
\clearpage
\appendix
\renewcommand{\thefootnote}{\arabic{footnote}}
\section{Storyboard Dataset}

\subsection{More Details about the Data Construction}

In this section, we provide a detailed description of the four stages in our storyboard data collection pipeline, along with the specific procedures applied at each stage.

\label{sec:data}
\noindent\textbf{Data Curation.} To ensure the quality of raw videos collected from online sources, we apply multiple filtering criteria. Specifically, we discard videos released before 2015, those with a resolution below 720p, or a bitrate lower than 800 kbps, retaining only clips with minimal motion artifacts and high spatial clarity.

\noindent\textbf{Scene Detect \& Keyframe Extract.} To extract high-quality keyframes from videos, we first segment each video into shots using PySceneDetect~\cite{PySceneDetect}. Within each shot, we compute the Laplacian score for all frames to assess sharpness and rank them accordingly, while also estimating optical flow to measure motion magnitude. Based on the combined rankings of sharpness and motion, we select the clearest and most distinct frames as keyframes.

\noindent\textbf{Quality Filter.} For each extracted keyframe, we apply multiple evaluation methods,including AES scoring~\cite{aes}, VLM-based assessment~\cite{guo2025seed1, xu2024visionreward}, and image quality metrics, to ensure visual fidelity. We additionally use watermark and subtitle detection tools~\cite{watermark} to filter out any keyframes containing such artifacts.

\begin{figure}[t]
  \centering
   \includegraphics[width=0.9\linewidth]{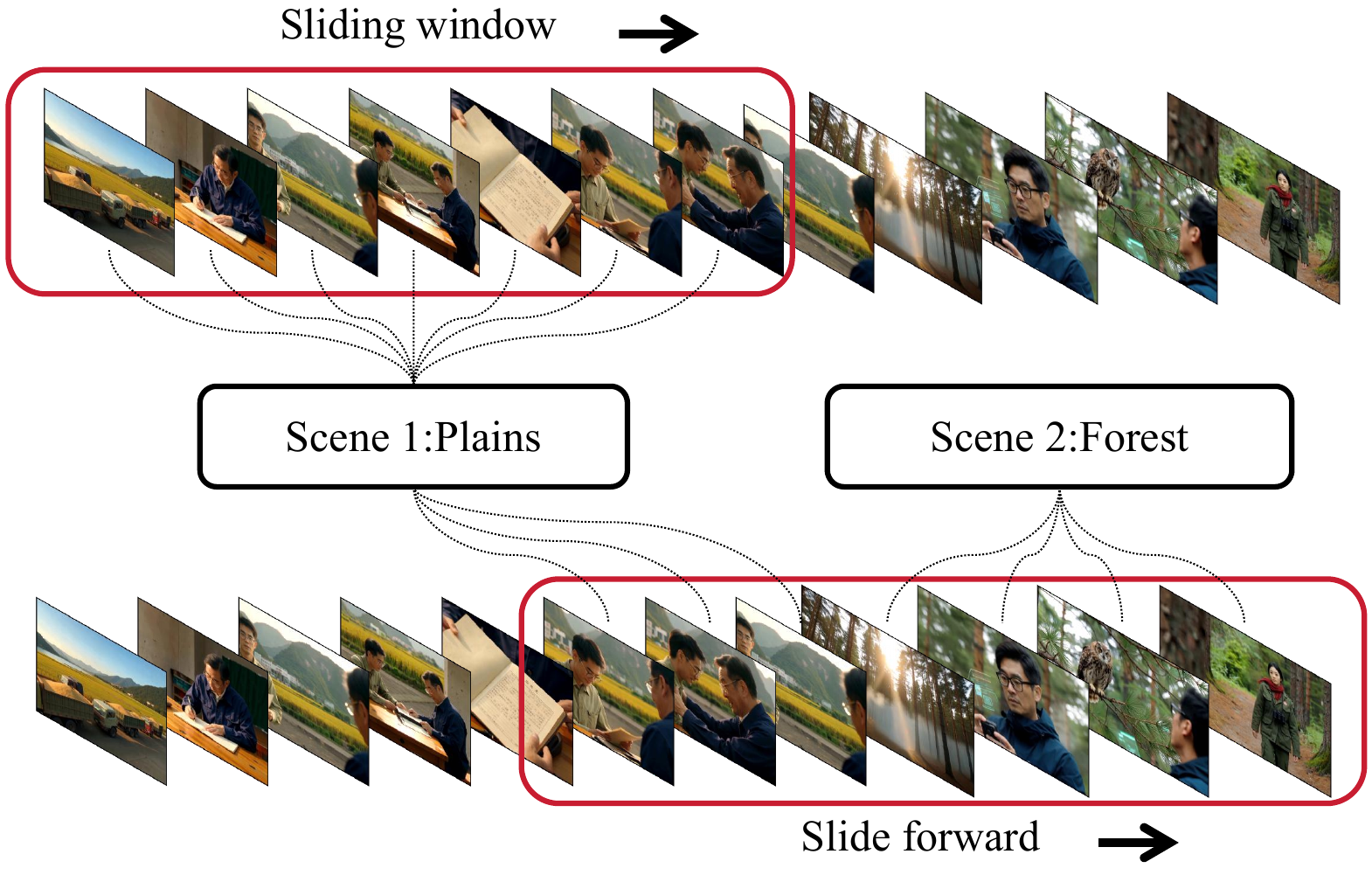}
   \caption{Using a sliding-window strategy with a VLM to extract storyboard groups from the same scene.}
   \label{fig:slide window}
\end{figure}

\noindent\textbf{Scene-Wise Grouping.} After extracting keyframes, we group those belonging to the same scene and sharing narrative continuity. Since feeding too many frames into a VLM can degrade its accuracy, we adopt a sliding-window strategy, as shown in \cref{fig:slide window}: keyframes are processed in small batches, and the VLM identifies those that form a coherent narrative sequence within each scene. Overlapping frames between windows are then used to determine the storyboard group to which each keyframe belongs, ensuring both temporal order and grouping accuracy.

\noindent\textbf{Role Extraction.} When extracting roles from each storyboard group, roles often appear repeatedly across shots. Directly applying an instance segmentation model to each shot would therefore produce many duplicate detections. To address this, we further use a VLM to merge and deduplicate the extracted character regions. We avoid face-embedding–based clustering~\cite{deng2019arcface}, as face recognition methods are unreliable for animated content or non-human characters.

\begin{figure*}[t]
  \centering
   \includegraphics[width=0.99\linewidth]{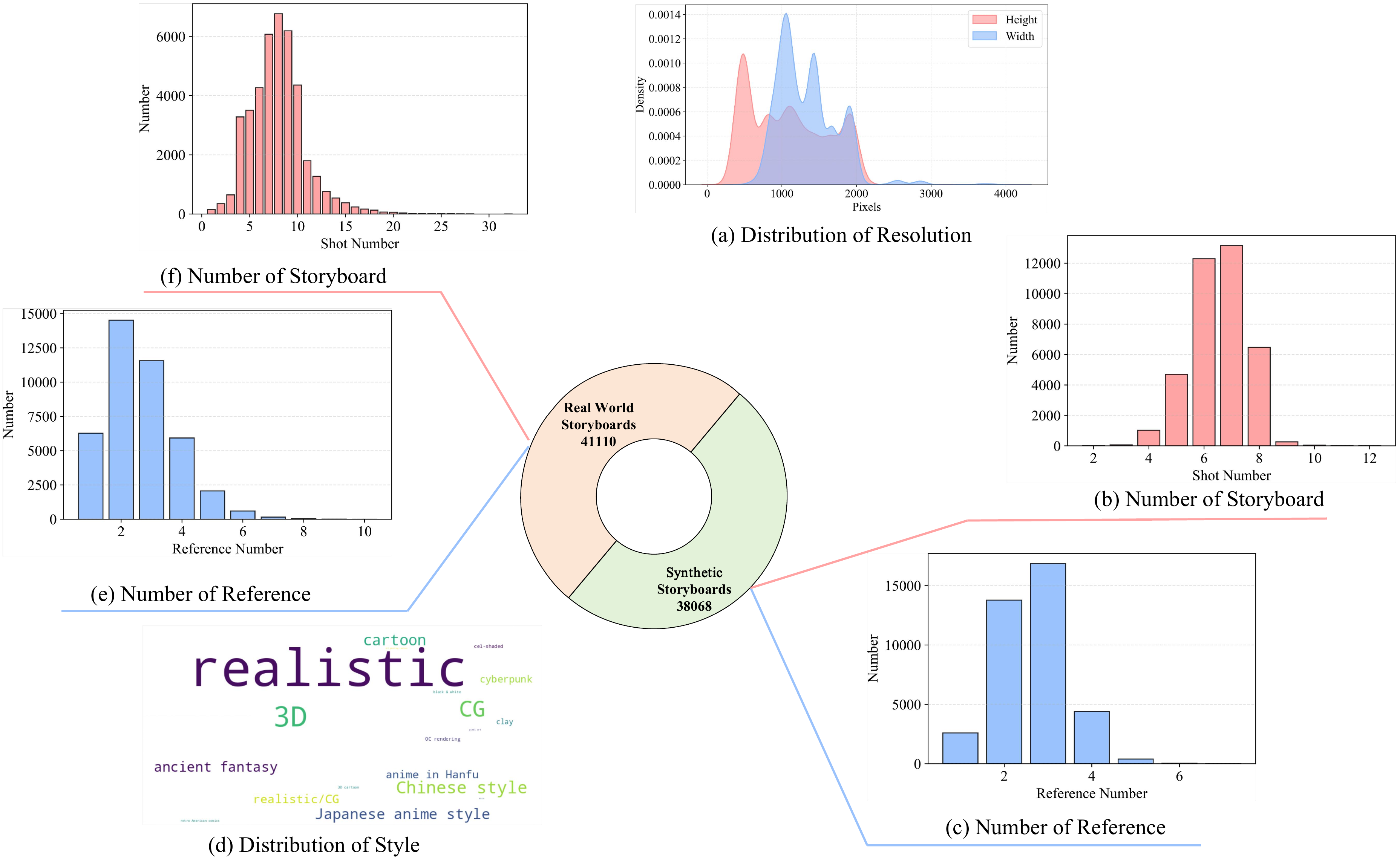}

   \caption{Data statistics of the DreamShot storyboard dataset. We report statistics for both data sources (real videos and AIGC-generated videos), including: the total number of storyboard samples; (a) resolution distribution of all samples; (b) shot-count distribution for synthetic storyboards; (c) reference-count distribution for synthetic storyboards; (d) overall style distribution; (e) reference-count distribution for real storyboards; and (f) shot-count distribution for real storyboards.}
   \label{fig:data analysis}
\end{figure*}

\noindent\textbf{Multi-View Role Generation.} Directly using roles extracted from the current shot as reference images can lead to severe copy–paste artifacts, causing the model to simply replicate the reference appearance. To increase reference diversity, we first apply cross-scene matching to retrieve the same characters from other scenes and use them as additional references. We further employ character augmentation techniques—such as rotating characters with video models~\cite{gao2025seedance} to obtain front, side, and half-body views, or generating diverse portraits with reference-image models~\cite{seedream2025seedream}. These generated candidates are further filtered using ArcFace~\cite{deng2019arcface} identity comparison to ensure that only images depicting the same role are retained. These augmented references provide richer appearance variations and effectively mitigate copy–paste behavior during generation.

\noindent\textbf{Story Annotation.} To ensure the narrative quality of the resulting storyboards, we use a VLM to generate structured annotations for all retained high-quality shots. The VLM provides shot-level descriptions covering perspective, environment, character actions, and stylistic attributes. For each reference roles, we further annotate detailed identity information, including gender, ethnicity, appearance, hairstyle, and clothing.

\noindent\textbf{Final Quality Control.}
For the final storyboard data, we perform an additional quality check using both a VLM and human review to assess narrative coherence, annotation accuracy, and visual clarity.

\subsection{Data Statistics}
Through the above pipeline, we obtain approximately 41K real-world storyboard samples and 38K AI-generated samples. We conduct a detailed analysis of the dataset, as shown in \cref{fig:data analysis}, including the distribution of shot counts, reference-role counts, and resolutions. Real-world storyboards exhibit a broader range of shot lengths, from 2 to 30 shots, with most falling between 5 and 12, while AI-generated storyboards are primarily concentrated between 4 and 8 shots. Similarly, most storyboard groups contain 2 to 4 reference characters. All samples maintain high visual resolution, with many reaching up to 2K. As shown in \cref{fig:data analysis} (d), the dataset also covers a wide variety of visual styles, including realistic, 3D, and anime aesthetics.

Based on this diverse dataset, we train our DreamShot model leveraging video-model priors to fully unlock its storyboard generation capability, achieving more consistent and more accurate shot synthesis.

\begin{table}[t]
    \centering
    \footnotesize
    \setlength{\tabcolsep}{0.5mm}{
    \begin{tabular}{lcccccccc}
    \toprule
        \multirow{2}*{DreamShot} & \multicolumn{2}{c}{CIDS(Character)} && \multicolumn{2}{c}{CSD(Style)} & AES & Alignment  \\
        \cline{2-3} \cline{5-6}
        \rule{0pt}{3ex} ~ & Self$\uparrow$ & Cross$\uparrow$  & & Self$\uparrow$  & Cross$\uparrow$  & Score$\uparrow$  & Score$\uparrow$  \\
    \midrule 
        \rowcolor{palered} 
        Dataset source & \multicolumn{7}{c}{Reference-to-Shot} \\
        w/ Real Data & 48.6 & \textbf{52.2} & & \textbf{64.7} & 44.6 & 5.24 & 3.33 \\
        w/ Synthetic Data & 46.4 & 49.3 & & 63.2 & 43.5 & \textbf{5.57} & 3.38 \\
         \rowcolor{mygray}
        Ours & \textbf{48.7} & 51.6 && 64.5 & \textbf{45.1} & 5.50 & \textbf{3.39}\\
    \bottomrule
    \end{tabular}
    }
    \caption{Effectiveness of Different Data Source.}
    \label{tab:data source}
\end{table}

\subsection{Effectiveness of Different Data Source}

\cref{tab:data source} summarizes the performance of models trained on different data sources. As shown, real storyboard data yield more consistent shot generation, particularly achieving a CDIS-Cross score of 52.2. This indicates that character identity and appearance in real storyboards align more faithfully with the reference characters, leading to improved cross-shot consistency.
In contrast, synthetic storyboard data achieve higher aesthetic scores and better alignment score, suggesting that their enhanced visual quality can further boost the fidelity of generated shots.
Therefore, combining real and synthetic data during training provides the best of both worlds—improving both the consistency and the overall quality of storyboard generation.

\begin{figure*}[t]
  \centering
   \includegraphics[width=0.6\linewidth]{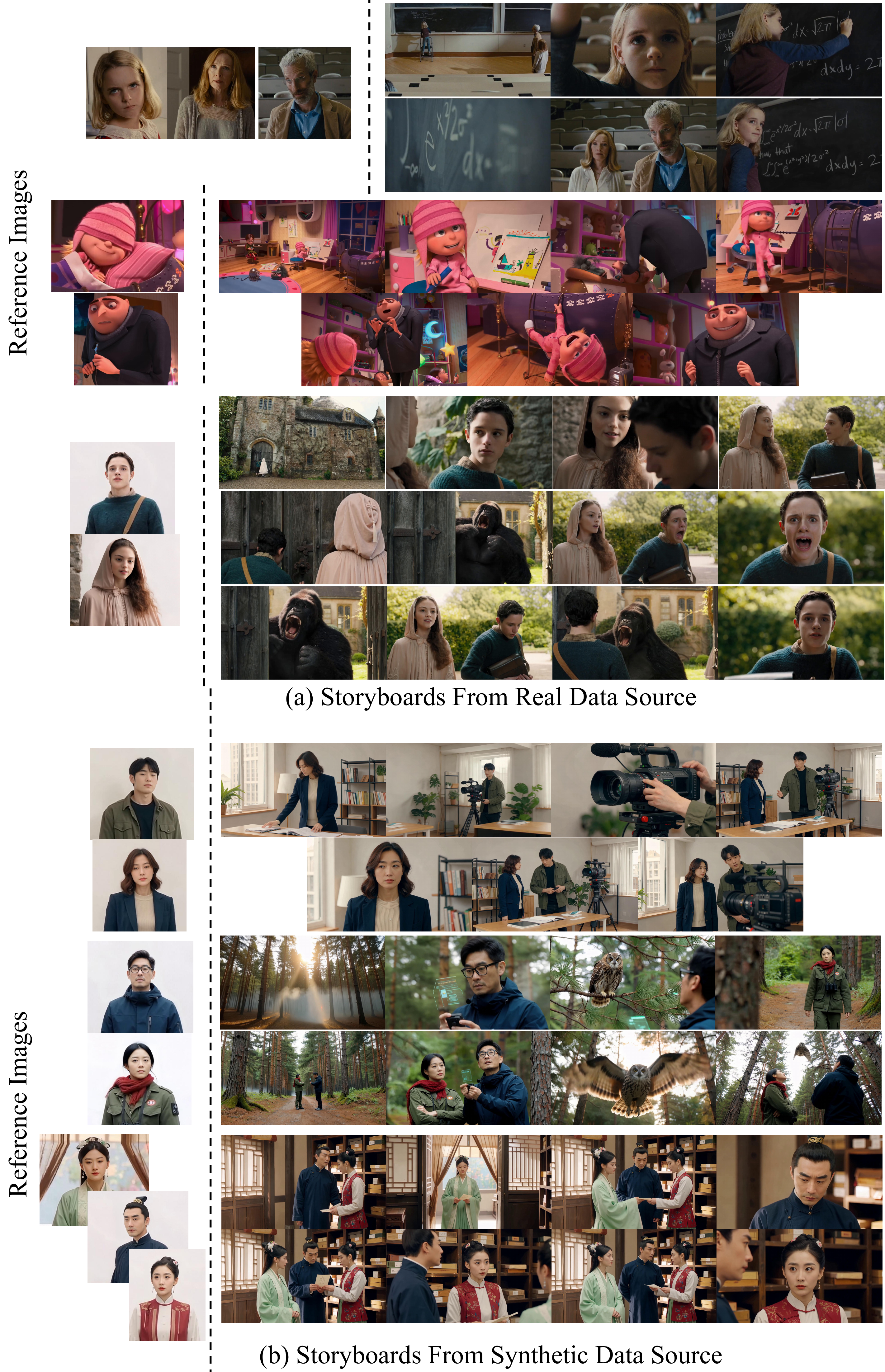}

   \caption{Storyboard samples from different data sources.}
   \label{fig:vis_dreamshot_dataset}
\end{figure*}

\cref{fig:vis_dreamshot_dataset} presents storyboard samples from different data sources. Real data exhibit stronger consistency and more expressive shot composition, although the overall aesthetic quality is sometimes lower and the lighting tends to be darker. In contrast, synthetic storyboard data show higher aesthetic quality and brighter, more visually appealing lighting conditions.

\section{Experiments}

\begin{figure}[t]
  \centering
   \includegraphics[width=0.99\linewidth]{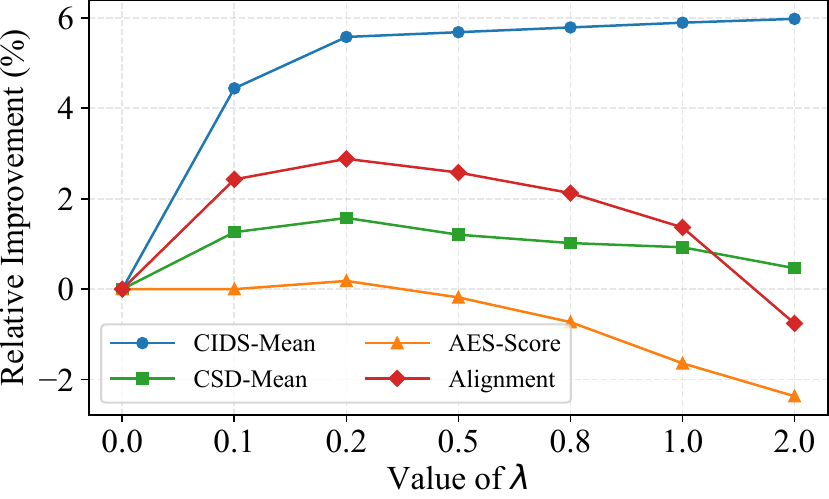}

   \caption{Effectiveness of $\lambda$ in Training Objective.}
   \label{fig:lambda}
\end{figure}

\subsection{Effectiveness of $\lambda$ in Training Objective}
In \cref{fig:lambda}, we further investigate the effectiveness of the coefficient $\lambda$ in the training objective. The table reports the relative improvement compared to the setting without $L_{\text{RAC}}$ (i.e. $\lambda=0$). We observe that as $\lambda$ gradually increases, the CIDS score consistently improves, reaching up to a 6\% gain. However, when $\lambda$ becomes as large as 0.5, the aesthetic score starts to decline. This is mainly because the increasing weight of $L_{\text{RAC}}$ dilutes the influence of $L_{\text{diff}}$, introducing interference that ultimately reduces visual quality.
Considering both consistency and aesthetics, we select $\lambda=0.2$, which provides a noticeable CIDS improvement while maintaining high visual quality.

\subsection{Performance under Different Numbers of Storyboards}
\begin{figure}[t]
  \centering
  
  \begin{subfigure}{0.99\linewidth}
    \centering
    \includegraphics[width=\linewidth]{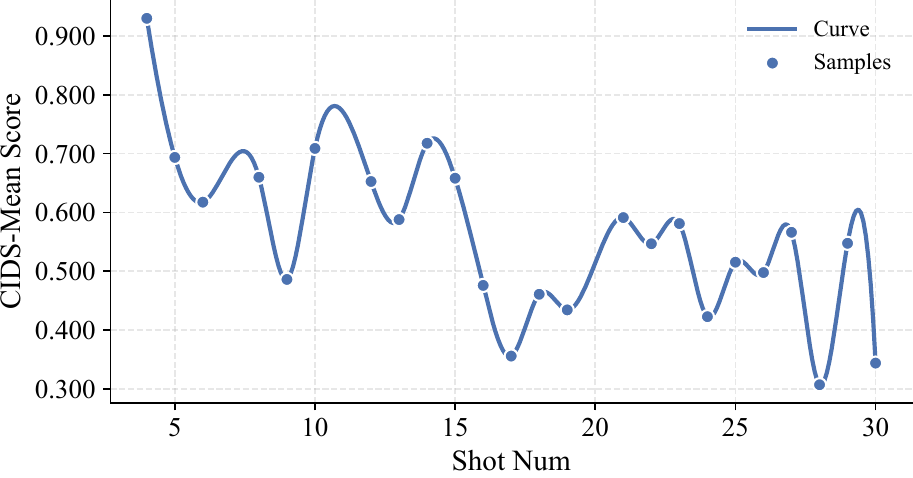}
    \caption{CIDS-Mean scores for different numbers of scenes.}
    \label{fig:shot_num_cids}
  \end{subfigure}
  \\[4pt]   

  \begin{subfigure}{0.99\linewidth}
    \centering
    \includegraphics[width=\linewidth]{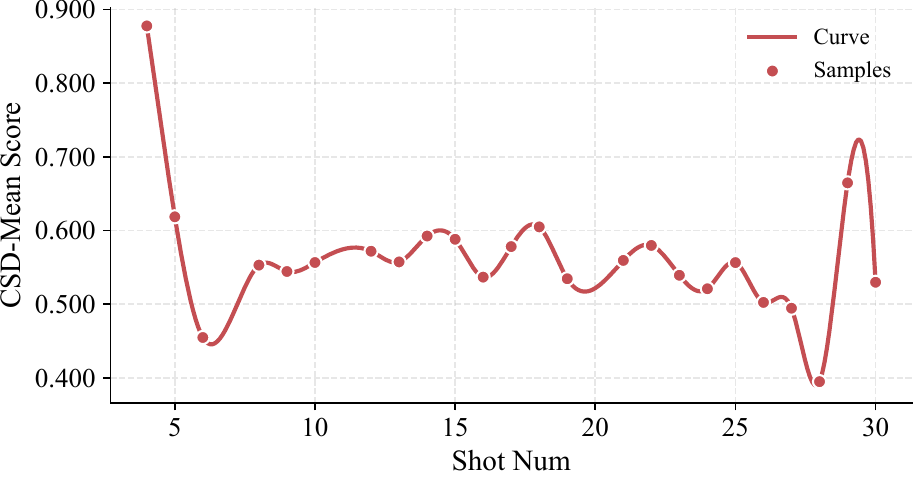}
    \caption{CSD-Mean scores for different numbers of scenes.}
    \label{fig:shot_num_csd}
  \end{subfigure}

  \caption{Effectiveness of Storyboard Number.}
  \label{fig:shot_num}
\end{figure}

\cref{fig:shot_num} illustrates the performance of DreamShot under different storyboard number. We observe that the CSD-Mean score remains largely stable across 4 to 30 shots, indicating that our method maintains strong scene-style consistency even in very long storyboards.
In contrast, the CIDS-Mean score gradually decreases as the number of shots increases, suggesting that character consistency becomes more challenging in ultra-long sequences. This limitation is primarily due to the distribution of our training data, where most storyboards contain only 6–10 shots.

\subsection{More Qualitative Results}

\begin{figure*}[t]
  \centering
   \includegraphics[width=0.99\linewidth]{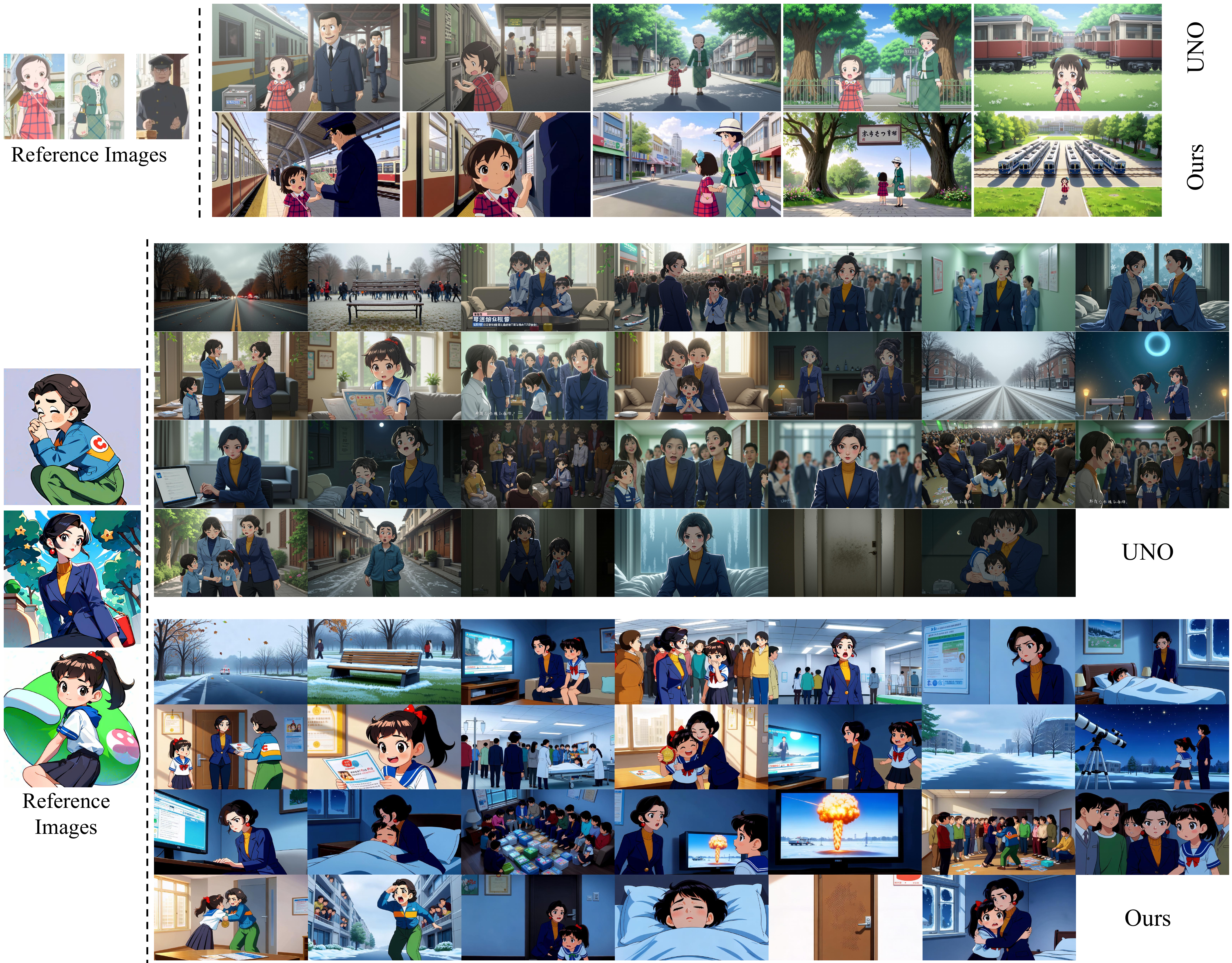}

   \caption{Qualitative results between our method and UNO~\cite{wu2025less} in VistoryBench~\cite{zhuang2025vistorybench}. Our method demonstrates superior character consistency, as reflected in the first storyboard group where fine-grained details such as the girl’s hair accessories are better preserved. In the case of long storyboard sequences, shown in the second group, our approach also exhibits strong generalization ability. The entire sequence maintains a more coherent style and scene layout compared with UNO.}
   \label{fig:vistory}
\end{figure*}

\noindent\textbf{Qualitative Results in VistoryBench.} \cref{fig:vistory} presents the quantitative comparison between our method and UNO~\cite{wu2025less} on VistoryBench~\cite{zhuang2025vistorybench}. VistoryBench primarily contains real reference images and each storyboard sequence may include multiple reference shots. The results show that our approach achieves strong performance in both character consistency and scene consistency.
Notably, when generating ultra-long storyboards, our method generalizes well despite the limited number of long-storyboard samples in the training data. The overall style and scene coherence remain consistently stable throughout the storyboards. This demonstrates both the strong generalization ability of our approach and its robustness in preserving character identity.

\begin{figure*}[t]
  \centering
   \includegraphics[width=0.99\linewidth]{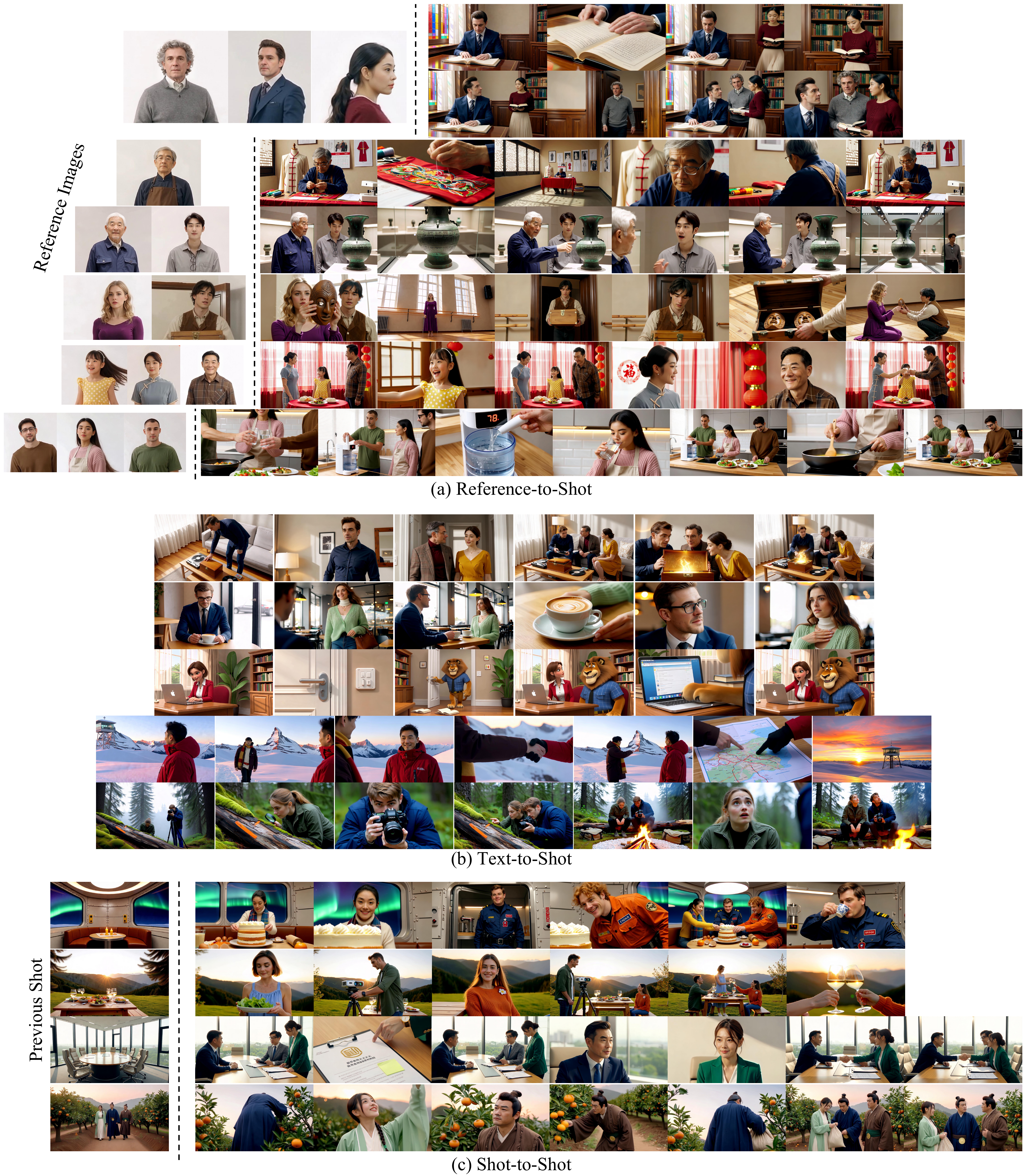}

   \caption{Quantitative results under Different Modes. (a) Reference-to-Shot Mode. Our method effectively preserves the character identity presented in the reference images. Moreover, for storyboards involving multiple reference roles, we demonstrates strong discriminative ability, with no noticeable confusion between different roles. (b) Text-to-Shot Mode. Our method responds well to prompts regarding shot transitions and viewpoint changes, while consistently maintaining scene coherence throughout the sequence. (c) Shot-to-Shot Mode. Our method maintains strong stylistic consistency with the previous shot and preserves the continuity of the appearing characters.}
   \label{fig:vis_dreamshot}
\end{figure*}

\noindent\textbf{Quantitative results under Different Modes.}
\cref{fig:vis_dreamshot} presents the quantitative results of our method under different generation modes. The results show that our approach performs well both in maintaining consistency with the reference images and in responding accurately to the shot-level textual descriptions. At the same time, DreamShot preserves strong scene coherence across shots, retaining the original environment even after camera transitions and supporting smooth narrative progression. 

\subsection{Storyboards to Long Video}
We further investigate DreamShot’s ability to support long-video creation through storyboard generation. Specifically, we employ a image-to-video (I2V) model~\cite{gao2025seedance} to convert each generated shot into a 5-second video clip, and then concatenate these clips to form videos longer than 30 seconds. The resulting long videos are provided in the supplementary folder.
The long video composed from our generated storyboards further demonstrates that the produced shots exhibit strong narrative coherence and consistent scene and character representation. Our method effectively decomposes long-form narratives into well-structured storyboard sequences, reducing the redundancy that often arises when generating long videos directly. Moreover, the storyboard representation offers clear advantages for editing and post-processing, as modifications to the final video can be achieved by editing individual shots rather than regenerating the entire sequence.

\section{Limitation}

\begin{figure}[t]
  \centering
   \includegraphics[width=0.99\linewidth]{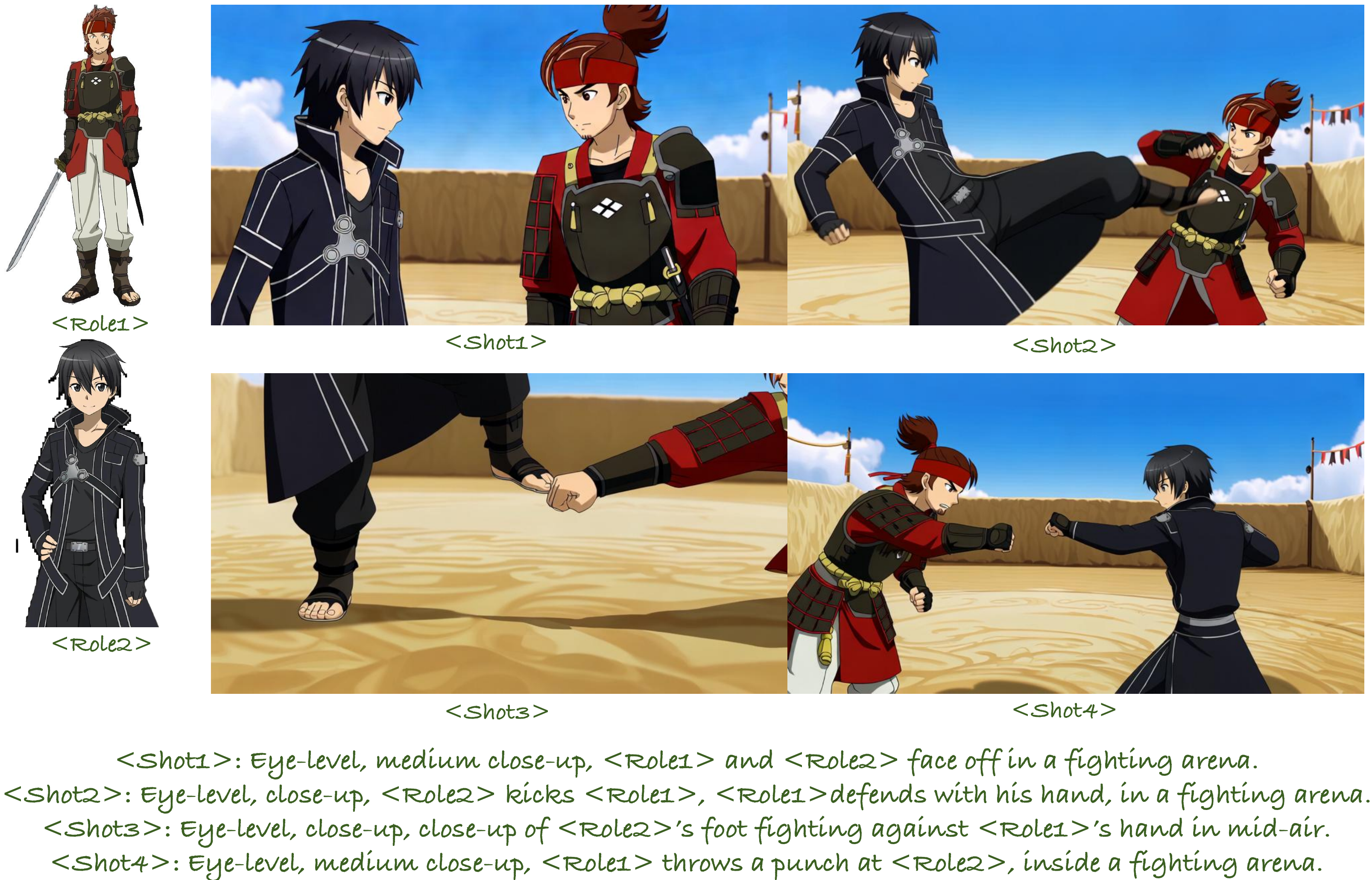}

   \caption{Examples of dynamic combat shots. DreamShot tends to generate clear and static storyboard frames, making it difficult to capture the sense of motion typically present in fighting storyboards.}
   \label{fig:vis_fight}
\end{figure}

\subsection{Performance under Extremely Long Storyboards}
In terms of storyboard number, DreamShot is primarily trained on storyboards containing around 6 to 10 shots. Although Dreamshot demonstrates strong generalization ability and can extend to longer sequences, its performance still degrades when facing ultra-long storyboards. As shown in \cref{fig:shot_num_cids}, the CIDS-Mean score gradually decreases as the number of shots increases. This limitation is largely attributed to the scarcity of long-storyboard samples in the training data.
In future work, we plan to explore strategies for improving character consistency in ultra-long storyboards. Possible directions include expanding data collection to incorporate more long storyboards, or adopting self-forcing or other autoregressive techniques to reduce error accumulation across extended shot sequences.

\subsection{Motion Storyboard}
Another limitation lies in the realism of action-oriented or combat-related shots. In such scenes, character motions are typically accompanied by dynamic blur, which conveys a stronger sense of movement, impact, and physical intensity. However, in cinematic data, it is often difficult to extract clean and representative frames from fast-motion sequences. Moreover, as discussed in \cref{sec:data}, our preprocessing pipeline employs a Laplacian-based filter to remove blurry frames. As a result, motion-blurred frames, which are essential for expressing dynamic actions, are frequently filtered out.
Consequently, most of the training samples become static shots, and DreamShot struggles to generate storyboard frames with strong dynamic motion cues or the visual intensity characteristic of combat scenes, as shown in \cref{fig:vis_fight}.

\subsection{Resolution}
Another limitation lies in the resolution of the generated outputs. This constraint primarily arises because existing video-based foundation models typically operate at around 480p or 720p, whereas image generation models can produce outputs at much higher resolutions, such as 1K. As a result, storyboards generated by video models may score lower on metrics such as aesthetic quality compared with those produced by high-resolution image models.
However, storyboard frames are often used as a guiding structure for downstream video production or as preliminary visual references for creators. In these scenarios, maintaining strong consistency across shots is more important and can provide more reliable creative guidance. In future work, we plan to explore higher-resolution storyboard generation to further enhance visual quality.

\end{document}